\documentclass[10pt,twocolumn,letterpaper]{article}

\usepackage{cvpr}              

\usepackage{mathtools}
\usepackage{amsmath}
\usepackage{graphicx}
\usepackage[inkscapelatex=false]{svg}
\usepackage{algorithm}
\usepackage{algpseudocode}
\usepackage{kotex}
\usepackage{multirow}
\usepackage{soul}
\usepackage{hyperref}
\usepackage{xcolor}

\definecolor{myred}{HTML}{B30000}
\definecolor{mygreen}{HTML}{006400}
\definecolor{myblue}{HTML}{0033A0}
\definecolor{myorange}{HTML}{FFA500}

\newcommand{\ysblue}[1]{\textcolor{myblue}{#1}}
\newcommand{\ysred}[1]{\textcolor{myred}{#1}}
\newcommand{\ysgreen}[1]{\textcolor{mygreen}{#1}}
\newcommand{\ysorange}[1]{\textcolor{myorange}{#1}}

\definecolor{cvprblue}{rgb}{0.21,0.49,0.74}

\def\confName{CVPR}
\def\confYear{2026}

\title{
HeSS: Head Sensitivity Score for Sparsity Redistribution in VGGT
}

\author{
Yongsung Kim$^{1}$~~Wooseok Song$^{2}$~~Jaihyun Lew$^{1}$~~Hun Hwangbo$^{1}$~~Jaehoon Lee$^{1}$~~Sungroh Yoon$^{1,2,3}$\textsuperscript{†}\\
$^1$IPAI, $^2$ECE, $^3$AIIS, ASRI, INMC, ISRC\\
Seoul National University\\
{\tt\small \{libary753, cody1129, fudojhl, genchiprofac, jhcaptain7, sryoon\}@snu.ac.kr}
}

\begin{document}
\maketitle
\let\thefootnote\relax\footnotetext{†Corresponding author}
\begin{abstract}
Visual Geometry Grounded Transformer (VGGT) has advanced 3D vision, yet its global attention layers suffer from quadratic computational costs that hinder scalability.
Several sparsification-based acceleration techniques have been proposed to alleviate this issue, but they often suffer from substantial accuracy degradation.
We hypothesize that the accuracy degradation stems from the heterogeneity in head-wise sparsification sensitivity, as the existing methods apply a uniform sparsity pattern across all heads.
Motivated by this hypothesis, we present a two-stage sparsification pipeline that effectively quantifies and exploits head-wise sparsification sensitivity. 
In the first stage, we measure head-wise sparsification sensitivity using a novel metric, the Head Sensitivity Score (HeSS), which approximates the Hessian with respect to two distinct error terms on a small calibration set. In the inference stage, we perform HeSS-Guided Sparsification, leveraging the pre-computed HeSS to reallocate the total attention budget—assigning denser attention to sensitive heads and sparser attention to more robust ones.
We demonstrate that HeSS effectively captures head-wise sparsification sensitivity and empirically confirm that attention heads in the global attention layers exhibit heterogeneous sensitivity characteristics. Extensive experiments further show that our method effectively mitigates performance degradation under high sparsity, demonstrating strong robustness across varying sparsification levels.
Code is available at \href{https://github.com/libary753/HeSS}{https://github.com/libary753/HeSS}.
\end{abstract}
\section{Introduction}
The task of multi-view 3D reconstruction is a long-standing problem in Computer Vision, and its performance greatly relies on the performance of Structure-from-Motion (SfM)~\cite{tomasi1992factorization, snavely2008internet, wu2013linearSFM, agarwal2009rome, schonberger2016colmap} and Multi-View Stereo (MVS)~\cite{Seitz2006EvaluationMVS, Goesele2007CommunityMVS, Furukawa2010MVS, Barnes2009PatchMatch}.
The recently introduced Visual Geometry Grounded Transformer (VGGT)~\cite{wang2025vggt} serves as a strong foundational model in this task, providing an end-to-end framework that unifies the traditionally separate tasks of SfM and MVS.
The success of VGGT stems from the core design of Global Attention (GA) and Frame Attention (FA) layers, stacked in an alternating manner.
The GA layer in particular, is crucial for understanding the overall scene structure, enabling comprehensive multi-view reasoning by capturing interactions between all frame tokens and inter-view relationships. However, this mechanism introduces substantial computational and memory overhead, as the cost scales quadratically with the number of input views.
This scalability issue limits VGGT’s practicality for large-scale or real-time 3D reconstruction scenarios.
 
\begin{figure}[t!]
    \centering
    \includegraphics[width=\linewidth]{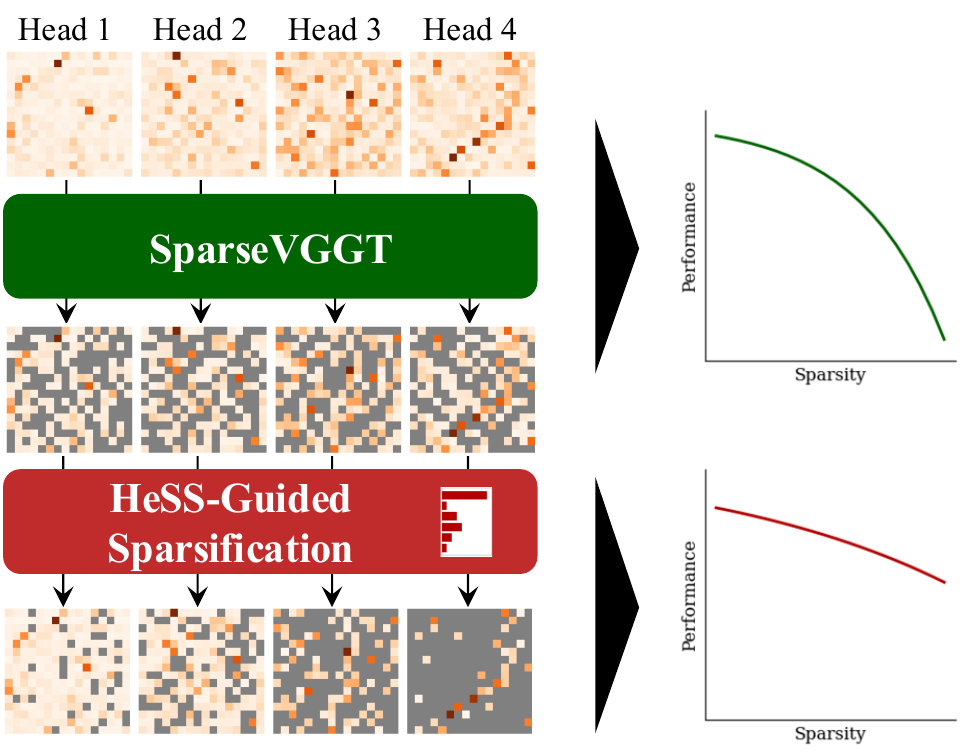}    
\caption{\textbf{Comparison with SparseVGGT~\cite{wang2025fastervggt}.} 
SparseVGGT suffers from severe performance degradation because it applies uniform sparsity across all heads, overly sparsifying sensitive ones. Our HeSS-Guided Sparsification preserves these highly sensitive heads, retaining performance even under high sparsity.
\vspace{-1.5em}
}
    \label{fig:teaser}
\end{figure}
To address this limitation of VGGT, several recent studies have attempted to reduce its computational cost, either by merging input tokens~\cite{shen2025fastvggt} or by applying sparse attention masks on the GA layer~\cite{wang2025fastervggt}. However, these approaches overlook the varying sensitivity of each attention head with respect to sparsified attention computations, leading to a significant drop in performance.
A more comprehensive discussion of related work is provided in the supplementary material.
We hypothesize that the performance degradation arises because existing methods impose a uniform sparsity pattern on all attention heads, despite their heterogeneous sparsification sensitivity.
Building on this idea, we propose a sparsification pipeline that adaptively assigns sparsity to each attention head based on head-wise sparsification sensitivity.
Our pipeline consists of two stages: a calibration stage, where we estimate the Head Sensitivity Score (HeSS) using a small calibration set, and an inference stage, where HeSS-guided sparsification is applied.

In the calibration stage, HeSS, the sensitivity of each attention head with respect to attention sparsification is computed using a calibration set and kept throughout the inference stage.
We define the sensitivity of each head as the Hessian of query projection weights in self-attention to two error terms: camera pose error and point cloud error.
The camera pose error serves as a signal for identifying attention heads that are critical for preserving the model’s understanding of the overall scene while the point cloud error compensates the camera pose error to preserve finer local geometry of each frame.

In the inference stage, we introduce a novel model sparsification strategy, named HeSS-Guided Sparsification, which masks out attention with head-wise sparsification ratios according to the calibrated HeSS.
This strategy reallocates attention resources across heads, applying stronger sparsification to robust heads and preserving more expressive attention for sensitive ones.

\begin{figure*}[ht!]
  \centering
   \includegraphics[width=\linewidth]{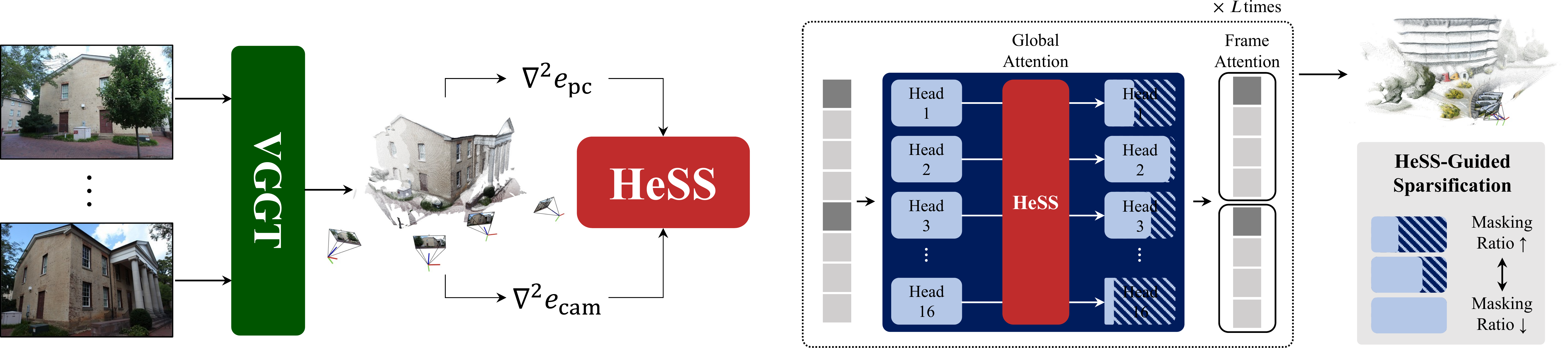}

   \caption{\textbf{Pipeline Overview.}
   Our pipeline consists of two stages: (a) Calibration stage, which computes Head Sensitivity Score (HeSS) of all attention heads in VGGT's Global Attention layers. The Hessian with respect to two errors, the camera pose error $e_\text{cam}$ and the point cloud error $e_\text{pc}$, is used to compute HeSS. HeSS is obtained from a calibration set, and these scores are fixed during the inference stage. (b) Inference stage, based on HeSS, different masking ratios are assigned throughout each attention head.
   }
   \label{fig:onecol}
\end{figure*}

Our main contributions are summarized as below:
\begin{itemize}
\item We find that attention heads exhibit different sensitivities to sparsification and propose \textbf{Head Sensitivity Score (HeSS)}, a novel metric to quantify this property.
\item We introduce \textbf{HeSS-Guided Sparsification}, a scheme that leverages HeSS to apply differentiated sparsification according to sensitivity, targeting robust heads more and sensitive heads less.
\item Our experiments demonstrate a superior performance-efficiency trade-off compared to existing approaches, validating our methodology.
\end{itemize}

\section{Preliminaries}

\paragraph{VGGT}
VGGT~\cite{wang2025vggt} is a unified multi-view transformer that takes a set of images $\{I_1, \dots, I_N\}$ as input and predicts their camera parameters (i.e., rotation matrices $R_i \in \mathbb{R}^{3 \times 3}$ and translation vectors $\mathbf{t}_i \in \mathbb{R}^3$), depth maps,
point maps,
and dense features for tracking.
For each image $I_i$, the image is patchified and the patch tokens are encoded using a pre-trained DINOv2 encoder~\cite{oquab2024dinov2} to be served as input tokens of VGGT.
In addition to the patch tokens, one camera token and four learnable register tokens are concatenated to form the initial input of VGGT.
The input tokens alternately go through global attention (GA) layers and frame attention (FA) layers in VGGT. As the name implies, the GA layer enables interaction of tokens across all frame tokens, which contributes in understanding of the overall structure of the scene, and the FA layer constrains attention within each frame for better structural understanding of each view.

\paragraph{Formulation of Multi-head Self-Attention (MHSA)}

Multi-head Self Attention (MHSA)~\cite{vaswani2017attention} is a very important mechanism in modern deep learning models. Here we aim to define the core formulation and notation of MHSA, as the GA layer in VGGT is also built upon this block.
Let $X\in \mathbb{R}^{S\times d}$ denote input tokens, where $S$ and $d$ are the number of tokens and dimensions, respectively.
MHSA employs attention on each head in parallel, where for each head $h$, $X$ is projected into query $Q^h\in \mathbb{R}^{S\times d_h}$, key $K^h\in \mathbb{R}^{S\times d_h}$ and value $V^h\in \mathbb{R}^{S\times d_h}$, where $d_h$ is the dimension of each head. 
The query, key and value are obtained by linear projection of $X$ using
projection matrices $W^h_Q\in\mathbb{R}^{d\times d_h}$, $W^h_K\in\mathbb{R}^{d\times d_h}$ and $W^h_V\in\mathbb{R}^{d\times d_h}$: $Q^h=XW^h_Q$, $K^h=XW^h_K$, $V^h=XW^h_V$.
Using the projected query, key and value, attention of a single head is computed as follows:
\begin{equation}
\operatorname{Attention}(Q^h, K^h, V^h)=\operatorname{softmax}\Big(\frac{Q^h(K^h)^\top}{\sqrt{d_h}}\Big)V
\end{equation}

\paragraph{SparseVGGT}

Despite VGGT's strong performance, the quadratic $\mathcal{O}(S^2)$ cost of self-attention becomes a severe bottleneck for large image sequences. In such settings, the activation footprint for $S$-token inputs—particularly the $S\times S$ attention maps—can easily surpass GPU memory capacity, necessitating a more efficient design.

To address this computational overhead, SparseVGGT~\cite{wang2025fastervggt} introduces a block-sparse attention mechanism for GA. This approach first computes a coarse approximation of the full attention map, and based on this approximation, the most significant regions, \emph{i.e.}, blocks, in the attention map are selected. Attention is then computed with unselected blocks masked out for efficiency.

In more detail, to coarsely identify important regions within the head's attention, SparseVGGT first applies pooling to $Q^h$ and $K^h$ and groups the queries and keys into smaller regions with size $b$. It then computes an approximated attention map $P^h = \operatorname{softmax}\left(\mathcal{P}_b(Q^h) \mathcal{P}_b(K^h)^\top \right)$, where $\mathcal{P}_b(\cdot)$ denotes average pooling function. We refer to each entry $P_{i,j}^h$  in this approximated attention map, as `attention blocks'.

SparseVGGT then sorts the attention blocks according to their probabilities $P_{i,j}^h$ in descending order.
Based on this order, blocks subject to evaluation are selected using two complementary criteria: a \textbf{CDF threshold ($\tau$)} and a \textbf{sparse ratio ($\rho$)}. The CDF threshold ensures that the selected blocks cover at least a $\tau$ fraction of the cumulative probability, while the sparse ratio acts as a lower bound, preserving a minimum of $k = \lfloor B \cdot (1-\rho)\rfloor$ top-ranked blocks. Based on this selection process, the \textit{attention budget} $c_h$ is defined as the number of activated attention blocks in head $h$. 
Identical hyperparameters $\tau$ and $\rho$ are used across all attention heads and layers, and thus the head-wise attention budgets $c_h$ are applied uniformly.
\section{Method}

\subsection{Method Overview}
In this section, we provide an overview of our overall method.
Our method builds upon SparseVGGT~\cite{wang2025fastervggt}, with a similar objective to predict and mask insignificant attention weights. 
To overcome the limitations of prior work that fail to account for the difference in head-wise sensitivity, we aim to reallocate the attention budget per head according to its sensitivity to attention weight masking—that is, in proportion to the performance degradation caused by attention masking. 

As shown in Fig.~\ref{fig:onecol}, our framework consists of two stages: the \emph{calibration} stage and the \emph{inference} stage.
In the calibration stage, we start by defining \textbf{He}ad \textbf{S}ensitivity \textbf{S}core (HeSS), with which we compute the sensitivity of each head with respect to attention weight masking.
HeSS is calculated using a small calibration set and is used throughout further model inferences. 
Based on this score, we demonstrate an effective method to distribute the attention budget by adopting denser attention weights in sensitive heads and sparser attention weights in less sensitive heads.
In Sec.~\ref{sec:hess}, we describe how HeSS is computed and in Sec.~\ref{sec:HeSS_masking}, we describe how the attention budget is distributed with HeSS during the inference stage.

\subsection{HeSS: Head Sensitivity Score}
\label{sec:hess}
We now introduce \textbf{He}ad \textbf{S}ensitivity \textbf{S}core (HeSS), a metric to quantify the sensitivity of each head in Global Attention (GA). 
This metric is designed to serve as guidance for a more effective sparsification strategy.

HeSS quantifies the sensitivity of model performance to the application of attention weight sparsification. Given that the primary objective is to mitigate performance degradation post-sparsification, our method is designed to identify attention heads that induce the least significant impact when masked. These low-sensitivity heads are thus the optimal targets for applying sparsity.
Specifically, model performance is assessed across two fundamental tasks, camera pose estimation and point cloud estimation, using their respective error metrics: $e_{\text{cam}}$ and $e_{\text{pc}}$.

Inspired by \cite{lecun1990obd}, we consider the curvature of error landscape as a proxy for evaluating sensitivity.
Accordingly, we use the Fisher Information Matrix (FIM), a commonly used tractable variant of the Hessian,  to measure head-wise sensitivity to attention masking. 
For each head, we combine sensitivity values derived from the error Hessians. More precisely, we approximate the Hessian of both $e_\text{cam}$ and $e_\text{pc}$ with respect to the query projection parameters, and these approximations are combined to obtain the final HeSS with weight $\lambda$.
The detailed formulation is described hereafter. 

\newcommand{\stopgrad}[1]{\operatorname{sg}\!\left(#1\right)} 

\paragraph{Camera Pose Error}

Camera pose estimation is one of the most crucial tasks in 3D vision, as it serves as the backbone for various downstream prediction tasks. 
As the coarse geometric scaffold of the entire scene, camera pose acts as a starting point on which all finer predictions depend.

For a model to accurately estimate camera poses across an image sequence, it must understand the geometric relationship among individual frames and their relative positions.     
To capture this importance at a holistic level, we define and utilize the camera pose error $e_\text{cam}$ as a primary performance metric.

To compute this error, we first obtain an optimal similarity transform matrix $\mathbf{H}$ by sequentially applying the Umeyama~\cite{umeyama2002least} and Iterative Closest Point (ICP)~\cite{besl1992icp} algorithms, to align the predicted and ground-truth point clouds.
To prevent gradients from flowing through the auxiliary alignment step, we apply stop-gradient constraint to $\mathbf{H}$.
With the alignment in place, we define $e_{\text{cam}}$ as the Mean Squared Error (MSE) between ground-truth camera positions ${\mathbf{t}}_i$ and the predicted camera positions $\hat{\mathbf{t}}_i$ transformed into the ground-truth coordinate system via $\mathbf{H}$:
\begin{equation}
\label{eq:loss_cam}
e_{\text{cam}}
= \frac{1}{2N} \sum_{i=1}^{N} \left| \operatorname{sg}(\mathbf{H})  \hat{\mathbf{t}}_i - {\mathbf{t}}_i \right|_2^2,
\end{equation}
where $\operatorname{sg}(\cdot)$ denotes the stop-gradient operator.

\paragraph{Point Cloud Error}
Although $e_\text{cam}$ successfully evaluates the model’s global geometric consistency, it offers limited competence in capturing fine structural accuracy. To complement this, we leverage VGGT's 3D scene reconstruction task.
This task requires the model to construct a dense 3D representation by predicting a per-pixel pointmap for every frame, fusing these predictions into a single point cloud in a unified coordinate space. To accomplish this successfully, the model must accurately capture fine-grained, pixel-level details and regress each pixel to its correct 3D location within the scene.
Consequently, the quality of the final reconstructed point cloud serves as a natural indicator of the model’s ability to capture these fine spatial details. In line with this idea, we define a point cloud error $e_\text{pc}$ to evaluate the model performance at this finer level.

Let $\hat P=\{\hat{\mathbf{p}}_j\}_{j=1}^{J}$ be the predicted point cloud and ${P}=\{{\mathbf{p}}_{k}\}_{k=1}^{K}$ the ground-truth point cloud. We compare predictions to the GT after applying $\operatorname{sg}(\mathbf{H})$, as in the camera pose error. We then define the inlier set with a confidence threshold $\epsilon$ from the predicted cloud as:
\begin{equation}
\label{eq:inlier_set}
\mathcal{I}
=\Big\{\, j\in\{1,\dots,J\}\ \Big|\ \min_{{\mathbf{p}}\in {P}}
\big\|\operatorname{sg}(\mathbf{H})\,\hat{\mathbf{p}}_j-{\mathbf{p}}\big\|_2^{2} < \epsilon \Big\},
\end{equation}
considering only high-confidence correspondences for the sake of robustness. We then define $e_{\text{pc}}$ as:
\begin{equation}
\label{eq:loss_pc_final}
e_{\text{pc}}
= \frac{1}{2|\mathcal{I}|}
\sum_{j\in\mathcal{I}}
\min_{{\mathbf{p}}\in {P}}
\big\|\operatorname{sg}(\mathbf{H})\,\hat{\mathbf{p}}_j - {\mathbf{p}}\big\|_2^{2}.
\end{equation}

\paragraph{HeSS Calculation}
With the two error terms $e_\text{cam}$ and $e_\text{pc}$ defined, we now describe how we compute the final sensitivity score for each head $h$. Our starting point is the Hessian of the error with respect to the query projection parameters $W_Q^h$.

Our rationale for focusing on these specific parameters is that they ($W_Q^h$ and $W_K^h$) fundamentally determine the head's behavior and its resulting attention logits. Thus, the head's overall impact on the loss is directly linked to the sensitivity of the loss with respect to these parameters.

Although both $W_Q^h$ and $W_K^h$ are viable candidates, we empirically find that using the Hessian with respect to $W_Q^h$ yields a more reliable head-wise sensitivity score than using the Hessian with respect to $W_K^h$. (Sec.~\ref{sec:ablation}) Therefore, we choose to use $W_Q^h$ in our approach.

However, computing the exact Hessian is computationally intractable. We therefore approximate it using Fisher Information Matrix (FIM) $\mathbf{F}$, which we estimate empirically from first-order gradients on a calibration dataset $\mathcal{D}$; additional details are given in the Appendix.

We denote the gradient with respect to $W_Q^{h}$ by $\mathbf{g}^{h}(\mathbf{x}) = \nabla_{W_Q^{h}} e(\mathbf{x})$. 
FIM for the camera pose and point cloud errors are then defined as:
\begin{align}
    \mathbf{F}_{\text{cam}}^{h} 
    &= \mathbb{E}_{\mathbf{x} \sim \mathcal{D}} 
       \left[ \mathbf{g}_{\text{cam}}^{h}(\mathbf{x}) 
              \left( \mathbf{g}_{\text{cam}}^{h}(\mathbf{x}) \right)^\top \right], \\
    \mathbf{F}_{\text{pc}}^{h} 
    &= \mathbb{E}_{\mathbf{x} \sim \mathcal{D}} 
       \left[ \mathbf{g}_{\text{pc}}^{h}(\mathbf{x}) 
              \left( \mathbf{g}_{\text{pc}}^{h}(\mathbf{x}) \right)^\top \right],
\end{align}
where, in practice, the expectations $\mathbb{E}_{\mathbf{x} \sim \mathcal{D}}[\cdot]$ are approximated by sample averages over $\mathcal{D}$. 

FIM captures the sensitivity of the camera pose and point cloud errors to the parameters of each head, but its magnitude can differ substantially across heads and between error types due to varying gradient scales. 
To obtain a single, comparable scalar sensitivity proxy for each head $h$, we compute the trace of its FIM and normalize it across all heads within the same GA layer:

\begin{align}
    \label{eq:HeSS_cam}
    \mathrm{HeSS}_{\text{cam}}(h) &=\frac{\operatorname{tr}(\mathbf{F}_{\text{cam}}^{h})}{\sum\nolimits_{h}\operatorname{tr}(\mathbf{F}_{\text{cam}}^{h})}, \\
    \label{eq:HeSS_pc}
    \mathrm{HeSS}_{\text{pc}}(h) &=\frac{\operatorname{tr}(\mathbf{F}_{\text{pc}}^{h})}{\sum\nolimits_{h}\operatorname{tr}(\mathbf{F}_{\text{pc}}^{h})}.
\end{align}

Then, we define the final HeSS as the weighted sum of the two component scores with weight $\lambda \in [0, 1]$:

\begin{equation}
\label{eq:HeSS_final}
\mathrm{HeSS}(h) = \lambda \mathrm{HeSS}_{\text{cam}}(h) + (1-\lambda)\mathrm{HeSS}_{\text{pc}}(h).
\end{equation}

Once the final HeSS is derived with a calibration dataset $\mathcal{D}$, it is kept fixed throughout the inference stage.
\begin{figure}[t!]
  \centering
   \includegraphics[width=\linewidth]{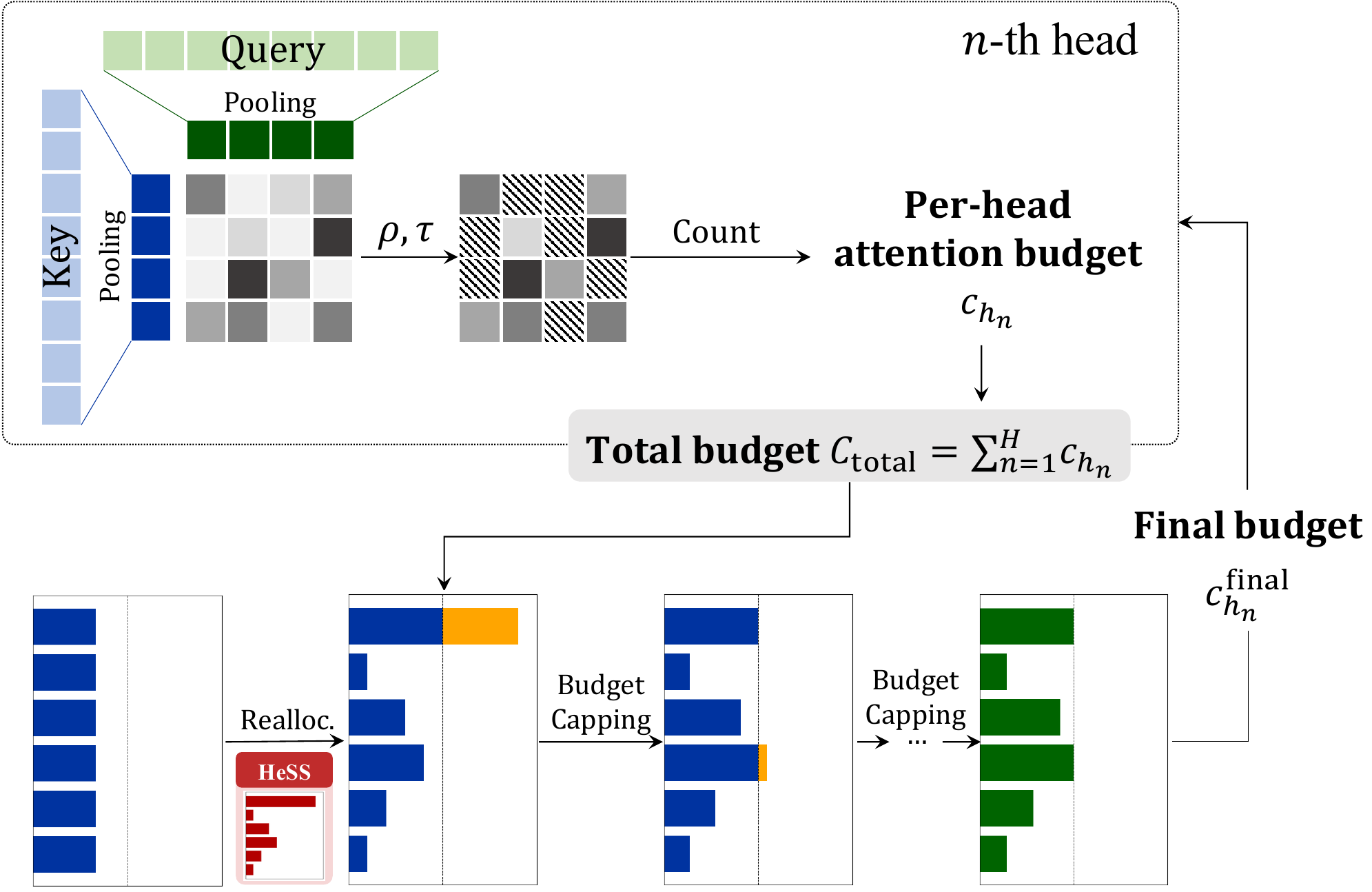}

    \caption{\textbf{HeSS-Guided Budget Reallocation.} The total budget ($C_{\text{total}}$) is obtained by summing baseline per-head budgets ($c_{h_n}$) (top). This budget is then reallocated based on HeSS scores (\ysred{red}). An iterative Budget Capping process redistributes surplus (\ysorange{orange}) from heads exceeding their capacity to the remaining uncapped heads, yielding the final budget ($c_{h_n}^{\text{final}}$, right).}
   \label{fig:hess_guided_budget_reallocation}
\end{figure}

\subsection{HeSS-Guided Sparsification}
\label{sec:HeSS_masking}

We now describe how HeSS is used during inference.
After the calibration stage produces head-wise HeSS, we keep these scores fixed and use them to guide sparsification at inference time. Our masking algorithm builds upon the block-sparse attention mechanism introduced in SparseVGGT~\cite{wang2025fastervggt}. As described in the preliminaries, SparseVGGT employs a
CDF threshold ($\tau$) and a sparse ratio ($\rho$) to reduce computational cost. Yet the same $\tau$ and $\rho$ criteria are applied uniformly to all heads, leaving head-wise sensitivity largely unexploited. 

In contrast, our \textbf{HeSS-Guided Sparsification} algorithm leverages HeSS to determine $c_h$ in a head-wise manner, allocating larger budgets to heads that are more sensitive to attention masking, and smaller budgets to heads that are less sensitive. This process consists of three main steps, as illustrated in~\cref{fig:hess_guided_budget_reallocation}.

\subsubsection{Total Budget Acquisition}
Our first step is to secure the total budget. Specifically, we sum the baseline per-head attention budgets to obtain the total budget, $C_{\text{total}} = \sum_{n=1}^{H} c_{h_n}$, where $H$ is the number of attention heads in each GA layer.
The baseline attention budget $c_{h_n}$ for each head are equal for all heads $h_n$ ($c_{h_1} =c_{h_2} = \dots = c_{h_H}$). 

\subsubsection{HeSS-Guided Budget Reallocation}
We reallocate the total budget $C_{\text{total}}$ in a prioritized manner
based on the calibrated HeSS. We compute a normalized reallocation weight $w_h$ for each head $h$:
\begin{equation}
w_{h} = \frac{\mathrm{HeSS}(h)}{\sum_{n=1}^H \mathrm{HeSS}(h_n)}.
\end{equation}
The ideal per-head budget is then given by $c_{h}' = C_{\text{total}} \cdot w_{h}$.

\subsubsection{Iterative Budget Capping}
This ideal budget $c_h'$ may be structurally infeasible, as a single dominant head could be allocated a budget that exceeds the total budget available to that head, $C_{\max}$. To solve this, we apply an iterative Water-Filling-style algorithm~\cite{gallager1968}. 

\begin{enumerate}
    \item \textbf{Cap:} Define the set of all overflowing heads $\mathcal{H}_\text{overflow} = \{h \mid c_h' > C_{\max}\}$. 
          Set their final budget to $c_h^{\text{final}} = C_{\max}, \forall h \in \mathcal{H}_\text{overflow}$.
    \item \textbf{Calculate Surplus:} Calculate the amount of total overflown budget 
          from these capped heads: $C_{\text{surplus}} = \sum_{h \in \mathcal{H}_\text{overflow}} (c_h' - C_{\max})$.
    \item \textbf{Reallocate:} Redistribute this $C_{\text{surplus}}$ \textit{only} 
          among the remaining (uncapped) heads ($\mathcal{H} \setminus \mathcal{H}_\text{overflow}$), 
          proportional to their original $\mathrm{HeSS}$ weights $w_h$.
    \item \textbf{Repeat:} Repeat steps 1-3 until $C_{\text{surplus}} \approx 0$.
\end{enumerate}

The final budget $c_h^{\text{final}}$ specifies the number of attention blocks each head selects from its approximate attention map, yielding a sparse attention mask shaped by head sensitivity.
Using the final budget $c_h^{\text{final}}$, each head selects a subset of blocks from its approximate attention map, resulting in a sparse attention mask informed by its sensitivity level.

\begin{figure*}[ht!]
    \centering
    \includegraphics[width=0.85\textwidth]{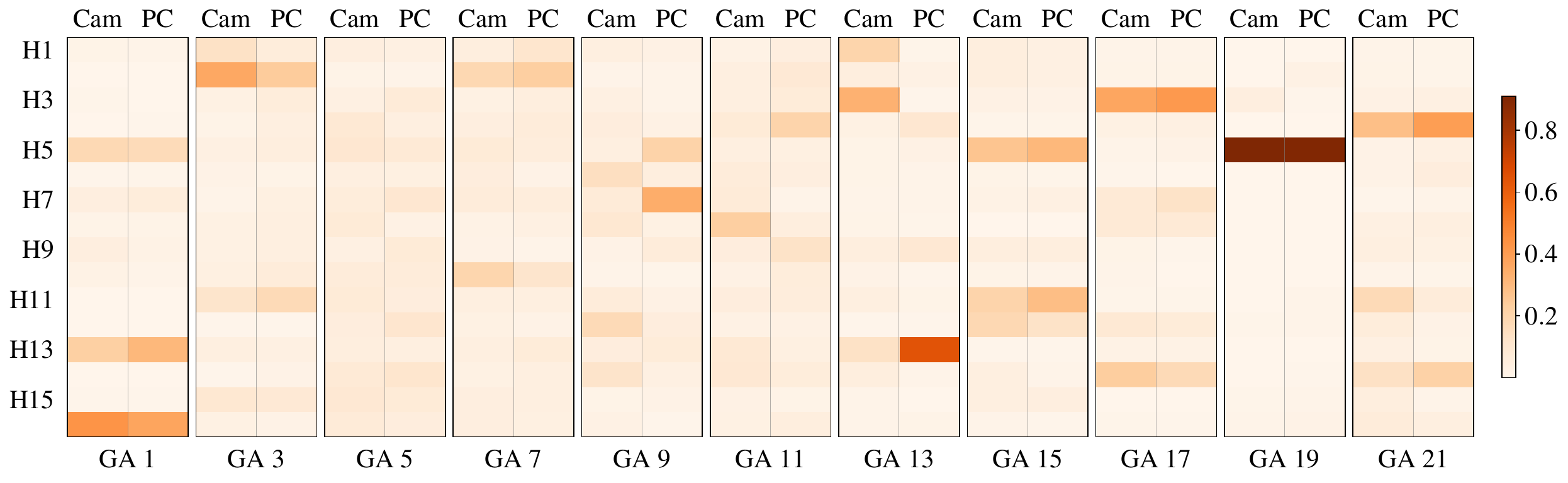}
    
    \caption{
        \textbf{HeSS Distribution in VGGT.}
The horizontal axis represents the Global Attention (GA) layers from GA 1 to GA 21. Each layer contains two columns, corresponding to $\mathrm{HeSS}_\text{cam}$ (Cam) and $\mathrm{HeSS}_\text{pc}$ (PC). The vertical axis lists the attention heads (H1–H16). Darker colors indicate higher sensitivity.
    }
    \label{fig:vis_scores_vggt}
\end{figure*}
\begin{figure*}[t!]
    \centering
    \includegraphics[width=\textwidth]{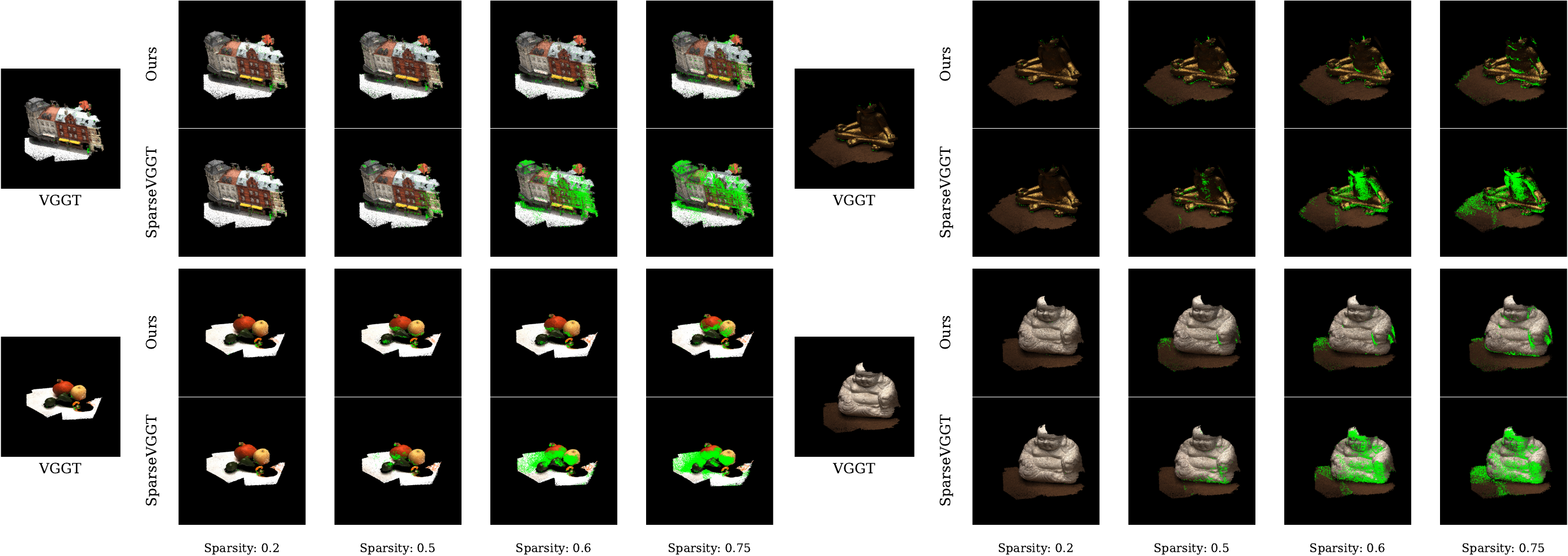}
    
    \caption{
        \textbf{Error visualization on DTU~\cite{jensen2014dtu}}
For each scene, we visualize predicted point clouds from VGGT~\cite{wang2025vggt}, SparseVGGT~\cite{wang2025fastervggt}, and our method at various sparsity levels.
Points whose 3D error exceeds 5 mm from the DTU ground truth are highlighted in green.
Our method produces fewer highlighted points than SparseVGGT, reflecting a more stable reconstruction as sparsity increases.
    }
    \label{fig:qual_res_1}
\end{figure*}
\begin{figure*}[t!]
    \centering
    \includegraphics[width=\textwidth]{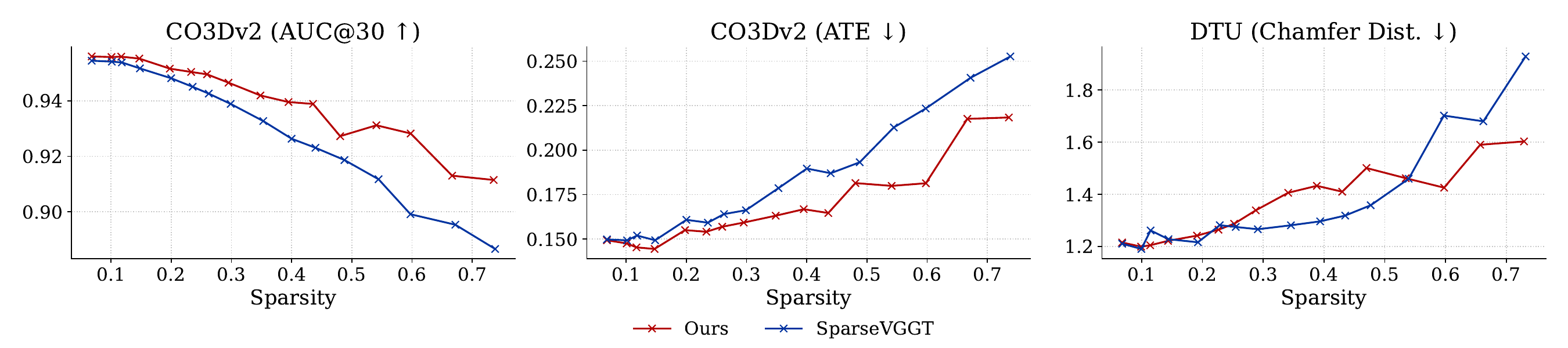}
    \vspace{-2em}
    \caption{\textbf{Comparison with SparseVGGT.} 
    We evaluate performance across various sparsity levels. On CO3Dv2~\cite{reizenstein2021co3d}, we measure camera pose estimation (AUC@30 $\uparrow$, ATE $\downarrow$). On DTU~\cite{jensen2014dtu}, we measure MVS performance (Chamfer Distance $\downarrow$).
    }
    \label{fig:performance_sparsity_trade_off}
\end{figure*}
\section{Experiments}

\subsection{Experimental Setup}
\paragraph{Implementation Details}
We set the inlier threshold to $\epsilon = 0.05$ and discard low-confidence points with confidence lower than 1 from VGGT outputs. 
For the experiments in this paper, we set $\lambda=0.5$ unless mentioned otherwise. 
To enable head-wise and block-wise attention masking, we employ a custom CUDA kernel as in SparseVGGT~\cite{wang2025fastervggt}. 
All experiments are conducted on NVIDIA RTX A40 GPUs.
For a fair comparison, we adopt the same $\tau$ and $\rho$ hyperparameters and do not mask out any attention weights involving camera and register tokens as in SparseVGGT~\cite{wang2025fastervggt}.
For example, the most aggressive masking configuration selects the top 20\% blocks and further includes those whose cumulative distribution (CDF) reaches up to 0.4, resulting in an overall sparsity of approximately 75\%.
We use the CO3Dv2 dev split for head-wise calibration, sampling 20 views per scene. 
Additional implementation details and hyperparameters are provided in the supplementary material.

\paragraph{Metrics}
We evaluate both camera pose estimation and MVS performance.
For camera pose estimation, we report AUC@30 following VGGT~\cite{wang2025vggt}.
Each AUC@30 measures the area under the cumulative distribution of the maximum of rotation and translation angular errors (in degrees) up to 30.
For MVS, we report Accuracy (prediction-to-ground-truth error) and Completeness (ground-truth-to-prediction error). Their mean corresponds to the symmetric Chamfer Distance.

\subsection{Analysis of Head Sensitivity Scores}
We visualize HeSS in~\cref{fig:vis_scores_vggt}. The results show a distinctly non-uniform distribution of sensitivity, with only a small subset of heads in each layer contributing substantially (e.g., H5 in GA 19).
This confirms that the sensitivity with respect to attention masking varies for each head.
Although many heads exhibit responses to both error terms, several display a clear task preference—for instance, H13 in GA 13 is more relevant to the camera pose loss, whereas H5 in GA 19 is strongly tied to the point cloud loss.

\subsection{Performance vs Sparsity Trade-off} 
We compare HeSS-Guided Sparsification against SparseVGGT~\cite{wang2025fastervggt} through both qualitative and quantitative evaluations.  
Qualitatively, in~\cref{fig:qual_res_1}, where points whose 3D error exceeds 5\,mm from the DTU ground truth are highlighted in green, our method shows fewer high-error regions, even under high sparsity, indicating that the reconstruction quality is well preserved.
Quantitatively, in~\cref{fig:performance_sparsity_trade_off}, our approach achieves consistently better performance on both camera pose estimation (CO3Dv2~\cite{reizenstein2021co3d}) and MVS (DTU~\cite{jensen2014dtu}), with a clear advantage in high-sparsity regimes.

\subsection{Comparison with ViT Sparsification.} 
\begin{figure}[t!]
    \centering
    \includegraphics[width=1\linewidth]{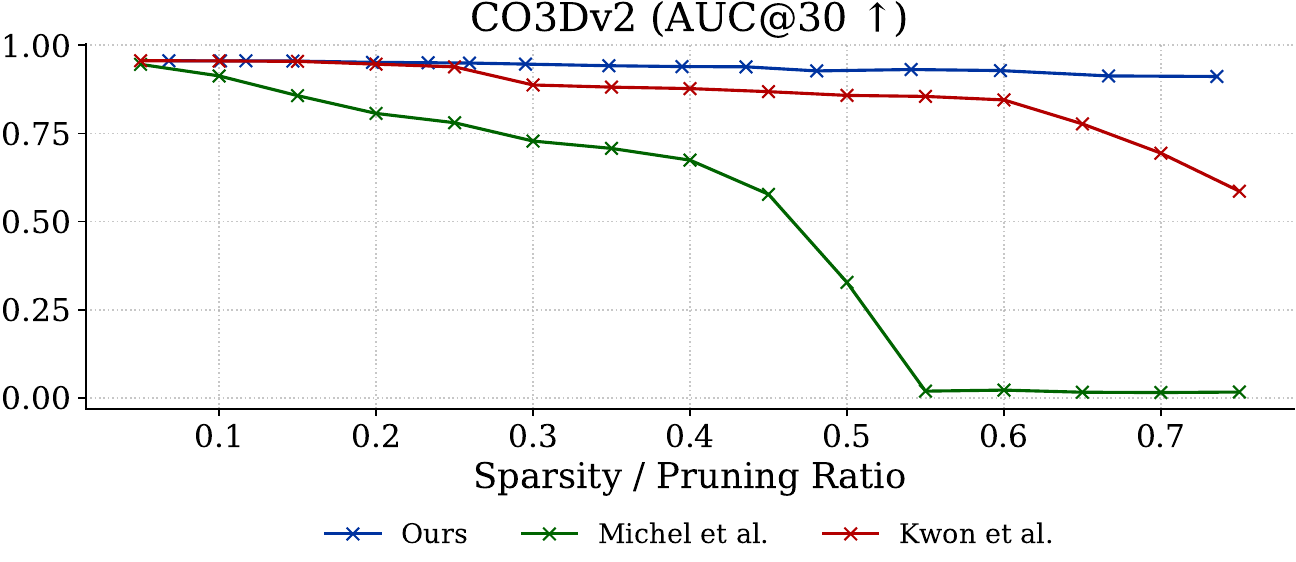}
    \caption{
    \textbf{Comparison with ViT~\cite{dosovitskiy2021vit} Sparsification.}
    Our method significantly outperforms baselines~\cite{michel2019sixteen, kwon2022fast} across all sparsity levels, maintaining robust performance even at extreme sparsity.
    }
    \label{fig:comp_w_sparsification}
\end{figure}

We compare our approach with head-level pruning~\cite{michel2019sixteen, kwon2022fast}. As shown in~\cref{fig:comp_w_sparsification}, our 3D-aware sparsification method significantly outperforms general sparsification baselines by a large margin.
While standard pruning is effective for 2D tasks, it lacks the geometric inductive bias needed for 3D reconstruction. Michel et al.~\cite{michel2019sixteen} suffers from rapid performance collapse beyond 50\% sparsity due to its coarse granularity. In contrast, our 3D-aware approach preserves the structural integrity of spatial tokens, maintaining high fidelity even at 75\% sparsity.

\subsection{Generalization}

\begin{figure}[t!]
    \centering
    \includegraphics[width=\linewidth]{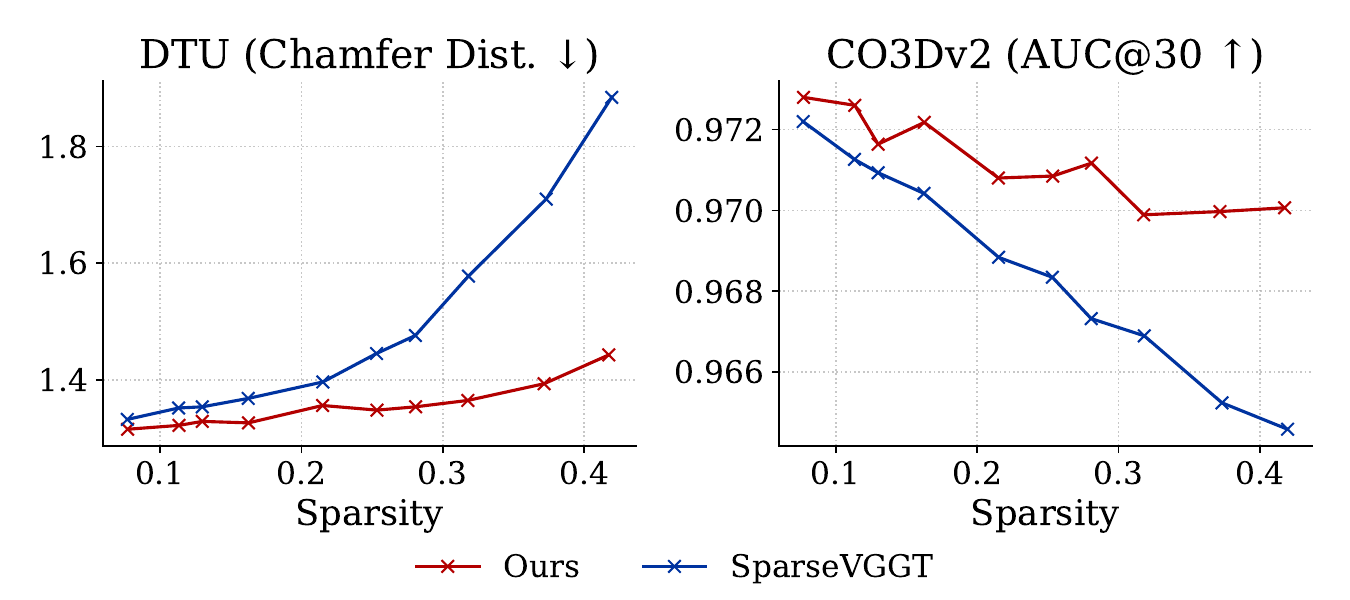}
    
    \caption{\textbf{Generalization to $\mathrm{\pi}^3$~\cite{wang2025pi}.} Performance comparison across sparsity levels on the $\mathrm{\pi}^3$. Our method shows improved robustness at higher sparsity levels compared to SparseVGGT~\cite{wang2025fastervggt}}
    \label{fig:performance_pi}
\end{figure}

To test the generalizability of our method, we apply our approach to the $\mathrm{\pi}^3$~\cite{wang2025pi} variant. 
We observe that $\mathrm{\pi}^3$ also exhibits non-uniform head sensitivity, as shown in~\cref{fig:vis_scores_pi3}.
We discover that $\mathrm{\pi}^3$ behaves slightly differently from VGGT, and that setting $\lambda=0$—using only $\mathrm{HeSS}_{\text{pc}}$—is sufficient and works best in practice.
The comparison with the baseline method is visualized in~\cref{fig:performance_pi}.
While SparseVGGT suffers from a severe performance drop, $\mathrm{HeSS}_{\text{pc}}$-guided method effectively preserves model performance.

\subsection{Sanity Check: Inverted HeSS ranking}
\begin{figure}[t!]
    \centering
    \includegraphics[width=\linewidth]{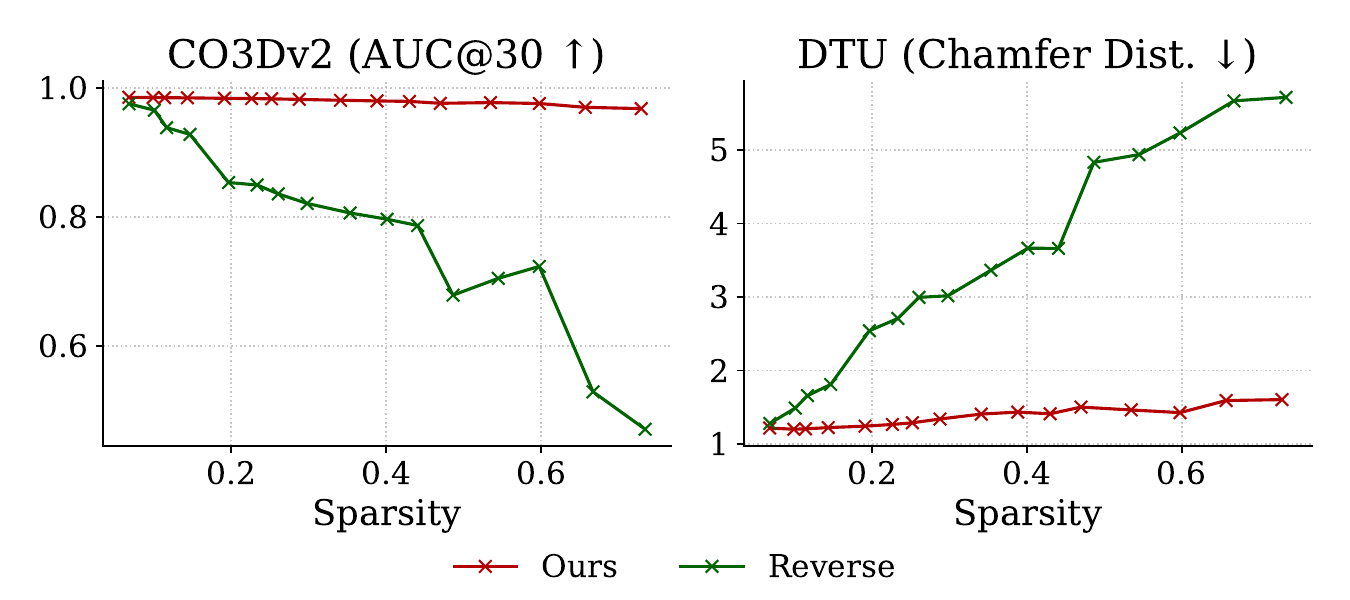} 
    
\caption{\textbf{Reversed HeSS Ranking.} 
Pruning sensitive heads (Reverse) triggers catastrophic failure, whereas HeSS maintains robust geometry by preserving high-impact heads.
\vspace{-1.0em}
}
    \label{fig:reverse_rank}
\end{figure}

We perform a sanity check by inverting the HeSS ranking. Given that HeSS is designed to identify and prune the least critical heads, the most sensitive heads are pruned first. As shown in~\cref{fig:reverse_rank}, this ``Reverse HeSS" strategy (green) causes a catastrophic performance collapse compared to our method (red), confirming the functionality of HeSS in correctly identifying critical heads.

\subsection{Ablation Study}
\begin{figure}[t]
    \centering
    \includegraphics[
    width=0.95\linewidth,
    ]{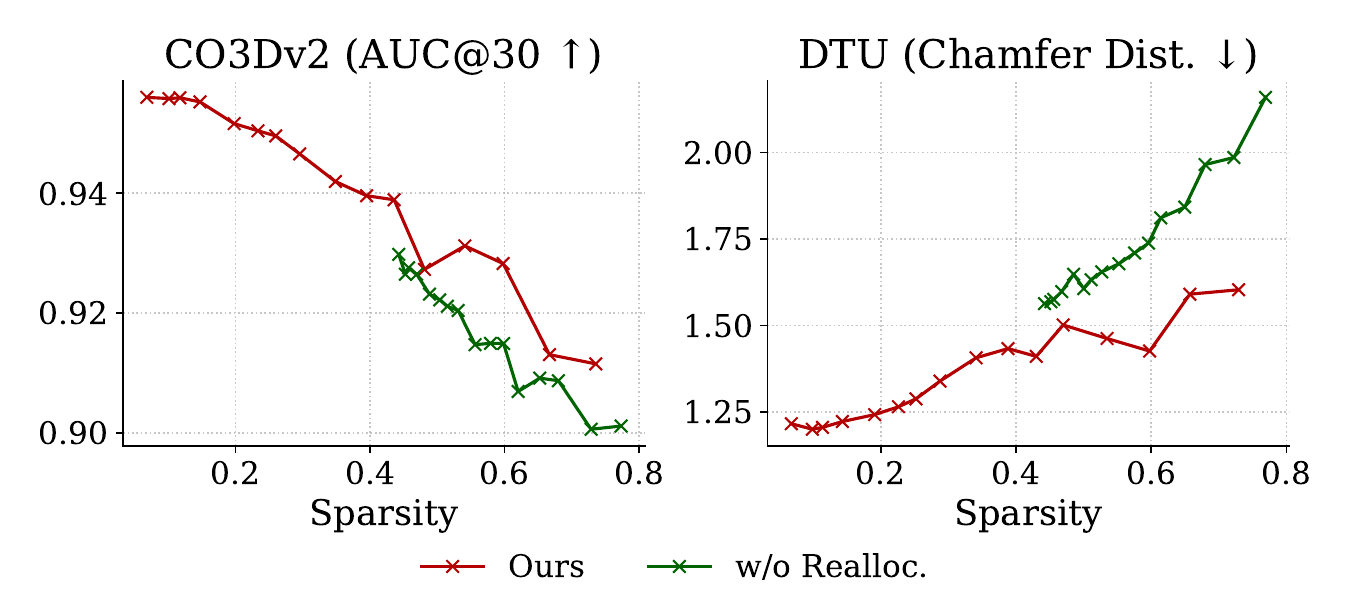}
    \caption{\textbf{Effect of budget reallocation.} Performance comparison with and without reallocation under the same sparsity budget. Without reallocation, surplus budgets remain unused, leading to degraded performance.}
    \label{fig:budget_realloc}
\end{figure}

\paragraph{Necessity of budget reallocation}
As shown in~\cref{fig:budget_realloc}, removing reallocation leads to unhandled surplus budgets and clear performance degradation, demonstrating that iterative budget capping is necessary beyond simple proportional allocation.


\paragraph{Projection Matrix Choice in HeSS}
\label{sec:ablation}
We evaluate the choice of the projection matrix in Hessian computation, using the DTU dataset. 
HeSS relies on FIM computed with respect to the query projection \(W^h_Q\). 
Because \(W^h_Q\) and \(W^h_K\) are the most directly connected parameters to attention logits and \(W^h_V\) is another crucial parameter in self-attention, we experiment on these three options.
Experiment on DTU shows that the choice of \(W^h_Q\) achieves stronger overall performance compared to \(W^h_K\) and \(W^h_V\).

\paragraph{Scaling}
We also evaluate different strategies for combining $\mathrm{HeSS}_{\text{cam}}$ and $\mathrm{HeSS}_{\text{pc}}$ on the DTU dataset. 
Each score is normalized by the sum of its FIM traces before averaging (\cref{eq:HeSS_cam,eq:HeSS_pc}), ensuring a balanced contribution between the two components.
We compare this sum-normalization to alternatives such as using raw trace values. 
As shown in~\cref{tab:ablation}, sum-normalization yields the strongest performance.

\begin{figure}[t!]
    \centering
    \includegraphics[width=1\linewidth]{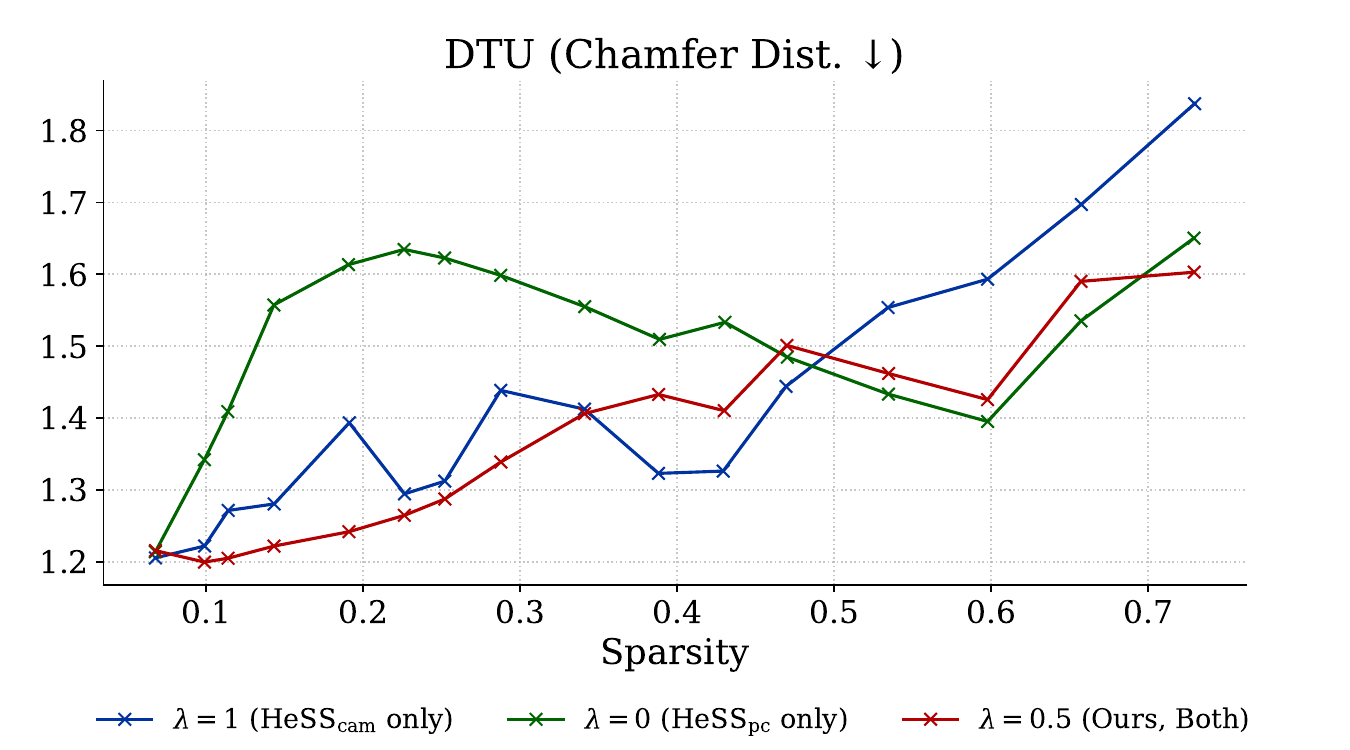}
    \caption{\textbf{Ablation of HeSS weight $\lambda$.}
    Performance comparison across sparsity levels for $\lambda\!=\!1$ (HeSS$_{\text{cam}}$ only, \ysblue{blue}), $\lambda\!=\!0$ (HeSS$_{\text{pc}}$ only, \ysgreen{green}), and $\lambda\!=\!0.5$ (Ours, \ysred{red}). Ours achieves the most stable performance across the entire sparsity range.
    \vspace{-0.5em}
    }
    \label{fig:hess_components}
\end{figure}

\paragraph{Ablation on HeSS weight $\lambda$.}
To validate our design of combining both error components, we conduct an ablation on $\lambda$:
setting $\lambda = 0$ corresponds to using only $\mathrm{HeSS}_{\text{pc}}$, while
$\lambda = 1$ corresponds to using only $\mathrm{HeSS}_{\text{cam}}$
in~\cref{fig:hess_components}. Using only $\mathrm{HeSS}_{\text{cam}}$ causes a performance drop at high sparsity, while using only $\mathrm{HeSS}_{\text{pc}}$ leads to performance degradation at low sparsity. Our full method, which combines both scores, maintains robust performance across the entire sparsity range, validating our combined approach. These results also clearly indicate the complementary nature of the two error components, where each addresses the shortcomings of the other at different sparsity levels.

\begin{table}[t]
\small
\centering
\begin{tabular}{c|ccc}
\toprule
 & Chamfer. $\downarrow$ & Acc. $\downarrow$ & Comp. $\downarrow$ \\ \midrule
$W^h_K$ & 1.9167 & 3.4498 & 0.3842 \\ 
$W^h_V$ & 1.9661 & 3.5400 & 0.3921 \\ 
\midrule
Linear  & \ul{1.8398} & \ul{3.2717} & 0.4078 \\  
Softmax & 1.9685 & 3.5536 & \ul{0.3834} \\ 
Log  & 2.1413 & 3.8816 & 0.4010 \\  
\midrule
Ours & \textbf{1.6030} & \textbf{2.8392} & \textbf{0.3667} \\ 
\bottomrule
\end{tabular}
\caption{
\textbf{
Ablation Study.
}
We compare different design choices for computing and combining \(\mathrm{HeSS}_{\text{cam}}\) and \(\mathrm{HeSS}_{\text{pc}}\). 
The top block evaluates the use of alternative projection matrices (\(W^h_K\) and \(W^h_V\)), while the bottom block compares combination strategies for \(\mathrm{HeSS}_{\text{cam}}\) and \(\mathrm{HeSS}_{\text{pc}}\). 
Our final configuration (Ours) achieves the best performance.
}
\label{tab:ablation}
\vspace{-1.5em} 
\end{table}

\subsection{Runtime Analysis}
HeSS-Guided Sparsification introduces only a negligible overhead, independent of input token length. For example, at 0.43 sparsity, our method's average per-scene runtime (measured over 10 runs across 50 view-scenes) in VGGT is 8.37 seconds, effectively matching the baseline SparseVGGT's 8.42 seconds. Despite the similar runtime, our approach achieves substantial performance gain compared to the baseline, as shown in \cref{fig:performance_sparsity_trade_off}.
\section{Limitations \& Future Work}
First, our pipeline treats all layers uniformly. However, different layers exhibit varying sensitivity to sparsification, and HeSS is not directly comparable across layers due to different Fisher Information scales. Developing layer-wise comparable sensitivity metrics could therefore be an important direction for future work.

Second, our approach focuses solely on inference-time sparsification and does not consider training-time adaptation. Designing models or training objectives that are intrinsically robust to sparsification—so that the network learns to maintain performance even under aggressive masking—would complement our method and present a promising avenue for improving the fundamental sparsifiability of large vision transformers.

\begin{table}[t!]
\small
\centering
\resizebox{0.85\linewidth}{!}{%
\begin{tabular}{cccc|c}
\toprule
Method & $\rho$& $\tau$& Sparsity & Runtime \\
\midrule
VGGT~\cite{wang2025vggt} & - & - & - & 10.35 \\
\midrule
SparseVGGT~\cite{wang2025fastervggt} & \multirow{2}{*}{0.5} & \multirow{2}{*}{0.9} & \multirow{2}{*}{0.43} & 8.42 \\
Ours & & & & 8.37 \\
\midrule
SparseVGGT~\cite{wang2025fastervggt} & \multirow{2}{*}{0.8}& \multirow{2}{*}{0.4} & \multirow{2}{*}{0.73} & 6.59 \\
Ours & & & & 6.58 \\
\bottomrule
\vspace{-1.5em}
\end{tabular}%
}
\caption{
\textbf{Runtime Analysis.} Comparison of inference runtime on an A40 GPU using the CO3Dv2~\cite{reizenstein2021co3d} dataset. At identical sparsity levels (0.43 and 0.73), HeSS-Guided Sparsification incurs negligible computational overhead.}
\label{tab:runtime}
\vspace{-1.5em}
\end{table}

\section{Conclusion}
In this paper, we investigate head-wise sensitivity in VGGT with respect to attention sparsification.
We propose HeSS, a metric to quantify this sensitivity, which is an approximated Hessian with respect to two error terms, camera pose error and point cloud error.
Moreover, we introduce HeSS-Guided Sparsification, a method that redistributes the total attention budget based on HeSS. Based on this redistributed attention budget, the sparsified attention operations display a robust performance.
Our experiments demonstrate the importance of handling attention heads differently based on their sensitivity.
This study underscores the value of analyzing the internal structure of VGGT and provides a principled pipeline for sparsifying it, which we expect to support future research on the efficiency and interpretability of multi-view models.

\section*{Acknowledgements}
This work was supported by Samsung Electronics Co., Ltd [No. IO250214-11971-01]; 
Information \& communications Technology Planning \& Evaluation (IITP) grant funded by the Korea government (MSIT) [No. RS-2021-II211343; RS-2022-II220959; RS-2025-02263754; 
Artificial Intelligence Graduate School Program (Seoul National University)]; 
the National Research Foundation of Korea (NRF) grant funded by MSIT [No. 2022R1A3B1077720; 2022R1A5A7083908] and the BK21 Four program of the Education and Research Program for Future ICT Pioneers, SNU in 2026.
This research was also conducted as part of the Sovereign AI Foundation Model Project (Data Track), organized by MSIT and supported by the National Information Society Agency (NIA) of Korea [No. 2025-AI Data-wi43].

{
    \small
    \bibliographystyle{ieeenat_fullname}
    \bibliography{main}
}

\clearpage
\setcounter{page}{1}
\setcounter{section}{0}
\renewcommand\thesection{\Alph{section}}
\setcounter{table}{0}
\renewcommand{\thetable}{S\arabic{table}}
\setcounter{figure}{0}
\renewcommand{\thefigure}{S\arabic{figure}}
\maketitlesupplementary

\section{Related Work}
\subsection{Learning-based MVS}
In traditional multi-view 3D reconstruction, Structure-from-Motion (SfM) reconstructs sparse 3D points and estimates camera poses from multi-view images.
Classical systems~\cite{tomasi1992factorization, snavely2008internet, wu2013linearSFM, agarwal2009rome, schonberger2016colmap} such as COLMAP~\cite{schonberger2016colmap} rely on hand-crafted modules, including keypoint matching~\cite{lowe2004sift} and bundle adjustment~\cite{triggs2000ba, wu2011mcba}. 
Multi-View Stereo (MVS)~\cite{Seitz2006EvaluationMVS, Goesele2007CommunityMVS, Furukawa2010MVS, Barnes2009PatchMatch} then densifies SfM by estimating per-view depth or dense point clouds. However, traditional pipelines are brittle in textureless regions, under occlusions and large viewpoint changes due to reliance on photometric consistency and geometric checks. With deep learning, learned features and matchers~\cite{detone2018superpoint, sarlin2020superglue} and learned MVS~\cite{Ji2017SurfaceNet, yao2018mvsnet, Yao2019RMVSNet} have substantially improved robustness and quality.

More recently, unified transformers~\cite{wang2024dust3r, leroy2024mast3r, wang2025cut3r, wang2025vggt, yang2025fast3r, wang2025pi} aim to replace the modular pipeline with end-to-end multi-view reasoning.
Pure self-attention models such as VGGT~\cite{wang2025vggt} and $\pi^3$~\cite{wang2025pi}, push this trend further by solving pose, depth and 3D structure within a single architecture, achieving strong accuracy and robustness.
However, self-attention over large numbers of image tokens remains computationally expensive, motivating methods that improve efficiency while preserving geometric fidelity.

\subsection{Efficient Vision Transformers}
VGGT is a recent Vision Transformer (ViT)~\cite{dosovitskiy2021vit}-based foundational model for 3D vision that processes many views jointly, making computational efficiency as critical as accuracy. However, since standard ViT blocks whose computational efficiency remains a well-known bottleneck, there already exists a large body of work devoted to improving ViT efficiency.

We categorize research on efficient ViTs into two lines: \emph{input-conditioned dynamic} and \emph{architecture-based}.  
Input-conditioned dynamic methods~\cite{rao2021dynamicvit, bolya2023tome, han2023dynamicperceiver, xu2023lgvit} analyze the model’s internal responses (e.g., activations and attention maps) to each input and make dynamic decisions such as patch or token pruning for efficiency.  
Architecture-based methods~\cite{yang2021nvit, yu2023xpruner, liu2023efficientvit, yang2024vitkd, yuan2022ptq4vit} improve efficiency via component-level analysis and architectural redesign, such as pruning unimportant heads, dimensions, or neurons~\cite{yang2021nvit, yu2023xpruner}, introducing memory-efficient modules~\cite{liu2023efficientvit}, applying knowledge distillation~\cite{yang2024vitkd}, and post-training quantization~\cite{yuan2022ptq4vit}.

Compared to single-image ViTs, VGGT’s multi-view design exacerbates these efficiency concerns. Recent VGGT-specific accelerations largely follow the input-conditioned dynamic paradigm: FastVGGT merges similar tokens without retraining~\cite{shen2025fastvggt}, and SparseVGGT replaces dense global attention with a block-sparse variant~\cite{wang2025fastervggt}.

\section{Additional Implementation Details}
\subsection{Geometric Alignment}
Geometric alignment is implemented using the Open3D library~\cite{zhou2018open3d}.
We adopt a coarse-to-fine registration strategy: an initial transformation is estimated via FPFH-based RANSAC, followed by precise refinement using the Point-to-Plane ICP algorithm.

\subsection{Nearest Neighbor Search Implementation}
For efficient nearest neighbor retrieval, we utilize the Faiss library~\cite{johnson2019billion} with an IVFFlat (Inverted File with Flat) index structure under the L2 metric. To ensure reproducibility, the index is trained deterministically on the CPU using K-means clustering with 4,096 centroids and subsequently transferred to the GPU for inference. 
Queries are executed with 16 probes to identify the single nearest neighbor. 
Finally, the Mean Squared Error (MSE) is computed for correspondences falling within a spatial tolerance of 0.05.

\begin{figure}[t]
    \centering
    \includegraphics[
    width=\linewidth,
    trim=0 0 0 12,
    clip
    ]{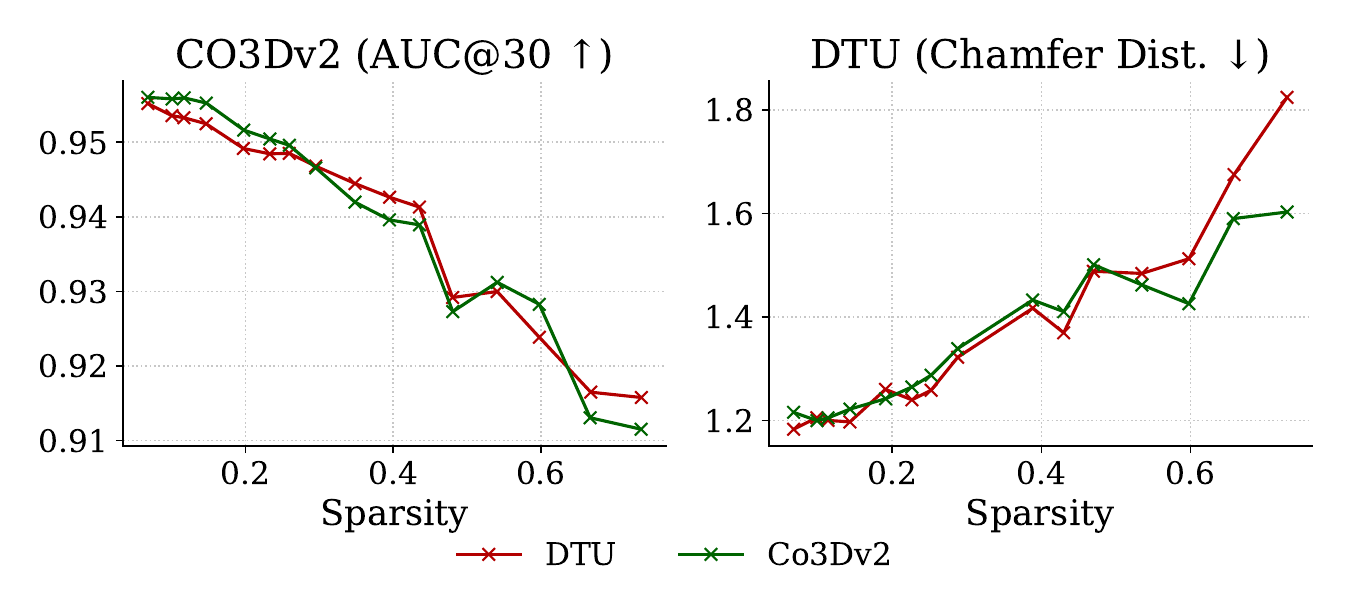}
    \caption{\textbf{Dataset sensitivity of HeSS calibration.} The performance remains largely unchanged regardless of the dataset used for calibration, indicating low sensitivity to the choice of calibration data.}
    \label{fig:calset_effect}
\end{figure}

\section{Calibration details}
We perform a one-time offline calibration using 40 scenes of CO3Dv2 (20 views each), which involves 80 backward passes (40 scenes $\times$ 2 error terms).
Regarding dataset sensitivity,~\cref{fig:calset_effect} demonstrates robustness: HeSS profiles from DTU and CO3Dv2 are interchangeable with minimal performance variance. This indicates our metric captures intrinsic model sensitivity rather than dataset specifics, though recalibration is recommended for extreme domain shifts.

\section{Derivation of the FIM from the Hessian}
In this section, we provide a detailed theoretical background and derivation for Head Sensitivity Score (HeSS) introduced in Sec. 3.2 of the main paper.

\subsection{Error Curvature as Sensitivity}
\label{sec:err_curv_as_sens}
We aim to quantify how each attention head $h$ influences the overall error.
Let $\mathbf{w}_h$ denote the vectorized query projection parameters of head $h$.
We assess its influence by analyzing how small perturbations in $\mathbf{w}_h$ affect the error function $e$ (corresponding to $e_{\text{cam}}$ or $e_{\text{pc}}$).
The second-order Taylor expansion around the current parameters gives:
\begin{multline}
e(\mathbf{w}_h + \Delta\mathbf{w}_h) \\
\approx e(\mathbf{w}_h) + \mathbf{g}_h^\top \Delta\mathbf{w}_h + \frac{1}{2} \Delta\mathbf{w}_h^\top \mathbf{H}_h \Delta\mathbf{w}_h .
\end{multline}
Here, $\mathbf{g}_h = \nabla_{\mathbf{w}_h} e(\mathbf{w}_h)$ denotes the gradient
and $\mathbf{H}_h = \nabla_{\mathbf{w}_h}^2 e(\mathbf{w}_h)$ denotes the Hessian,
both taken with respect to $\mathbf{w}_h$.

This expression shows that the perturbations in the parameters affect the error through
both gradient and curvature terms. 
However, as noted by LeCun et al.~\cite{lecun1990obd}, gradients vanish near a local optimum, making the first-order term negligible. 
Thus, the curvature term governed
by the Hessian $\mathbf{H}_h$ serves as the primary indicator of parameter
sensitivity. 

\subsection{FIM as a Hessian Approximation}
As established in~\cref{sec:err_curv_as_sens}, computing the full Hessian
$\mathbf{H}_h$ is computationally prohibitive. To clarify this point, we
derive the exact gradient $\mathbf{g}_h$ and Hessian $\mathbf{H}_h$ for
$e_{\text{cam}}$. Note that $e_{\text{pc}}$
(Eq. (4)) has the same mean-of-squares structure, so the
derivation applies analogously.

We begin with the error function $e_{\text{cam}}$ (Eq. (2)), where
$\mathbf{w}_h$ are the parameters of head $h$:

\begin{equation}
e_{\text{cam}}(\mathbf{w}_h) = \frac{1}{2N} \sum_{i=1}^{N} \left\| \mathbf{r}_i(\mathbf{w}_h) \right\|_2^2 = \frac{1}{2N} \sum_{i=1}^{N} \mathbf{r}_i^\top \mathbf{r}_i,
\end{equation}
where $\mathbf{r}_i(\mathbf{w}_h) = \operatorname{sg}(\mathbf{H}) \hat{\mathbf{t}}_i(\mathbf{w}_h) - {\mathbf{t}}_i$ is the residual vector for the $i$-th camera pose.
\paragraph{1. First Derivative (Gradient)}
\begin{align}
\mathbf{g}_h &= \nabla_{\mathbf{w}_h} e_{\text{cam}} = \frac{1}{2N} \sum_{i=1}^{N} \nabla_{\mathbf{w}_h} (\mathbf{r}_i^\top \mathbf{r}_i). \\
\mathbf{g}_h &= \frac{1}{2N} \sum_{i=1}^{N} 2 \left( \frac{\partial \mathbf{r}_i}{\partial \mathbf{w}_h} \right)^\top \mathbf{r}_i.
\end{align}
Letting $\mathbf{J}_i = \partial \mathbf{r}_i / \partial \mathbf{w}_h$ be the Jacobian matrix of the $i$-th residual with respect to the parameters $\mathbf{w}_h$, the gradient simplifies to:
\begin{align}
\mathbf{g}_h = \frac{1}{N} \sum_{i=1}^{N} \mathbf{J}_i^\top \mathbf{r}_i.
\end{align}
\paragraph{2. Second Derivative (Hessian)}
Differentiating the gradient yields:
\begin{equation}
\mathbf{H}_h
= \nabla_{\mathbf{w}_h} \mathbf{g}_h
= \frac{1}{N} \sum_{i=1}^{N}
\left(
\nabla_{\mathbf{w}_h}(\mathbf{J}_i^\top)\,\mathbf{r}_i
+
\mathbf{J}_i^\top \mathbf{J}_i
\right),
\end{equation}
where $\mathbf{J}_i = \partial \mathbf{r}_i / \partial \mathbf{w}_h$.
This gives the standard decomposition
\begin{equation}
\mathbf{H}_h
= \frac{1}{N} \sum_{i=1}^{N}
\left(
\underbrace{\mathbf{J}_i^\top \mathbf{J}_i}_{\text{Gauss–Newton}}
+
\underbrace{\sum_{k} r_{ik}\,\nabla_{\mathbf{w}_h}^2 r_{ik}}_{\text{Residual Hessian}}
\right).
\end{equation}
The second term requires the computation of the second-order derivatives of the residuals which is  computationally expensive. This difficulty motivates the use of the Empirical Fisher Information Matrix (FIM) as a surrogate for the Hessian. 

Under standard regularity conditions, when the error is defined as a negative log-likelihood term and the model fits the data reasonably well, the Fisher Information Matrix coincides with the expected Hessian of this error function~\cite{gentle_theory_stats}. In our case, the error terms are MSE losses, each of which can be interpreted as the negative log-likelihood of a Gaussian observation model. Since we assume that the target model is already well trained, we can leverage this connection and approximate the Hessian using the FIM. Concretely, we estimate the Empirical FIM from samples generated by the model and use this estimate as a tractable approximation to the true Hessian:
\begin{align}
\mathbf{F}_h &= \mathbb{E}_{\mathbf{x}\sim\mathcal{D}}\!\left[\mathbf{g}_h(\mathbf{x}) \mathbf{g}_h(\mathbf{x})^\top\right], \\
\hat{\mathbf{F}}_h &= {1 \over |\mathcal{C}|} \sum_{\mathbf{x}\in \mathcal{C}}\mathbf{g}_h(\mathbf{x}) \mathbf{g}_h(\mathbf{x})^\top,
\end{align}
where $\mathbf{g}_h(\mathbf{x})$ denotes the per-sample gradient of the error for input $\mathbf{x}$, $\mathcal{D}$ is the data distribution of $\mathbf{x}$, and $\mathcal{C}\subset\mathcal{D}$ is the calibration subset. Here, $\mathbf{F}_h$ denotes the FIM, and $\hat{\mathbf{F}}_h$ is its Empirical FIM computed over the calibration set $\mathcal{C}$; both matrices 
rely only on the first-order gradients and serve as a tractable approximation to the true curvature.

\begin{figure*}[t!]
    \centering
    \includegraphics[width=\textwidth]{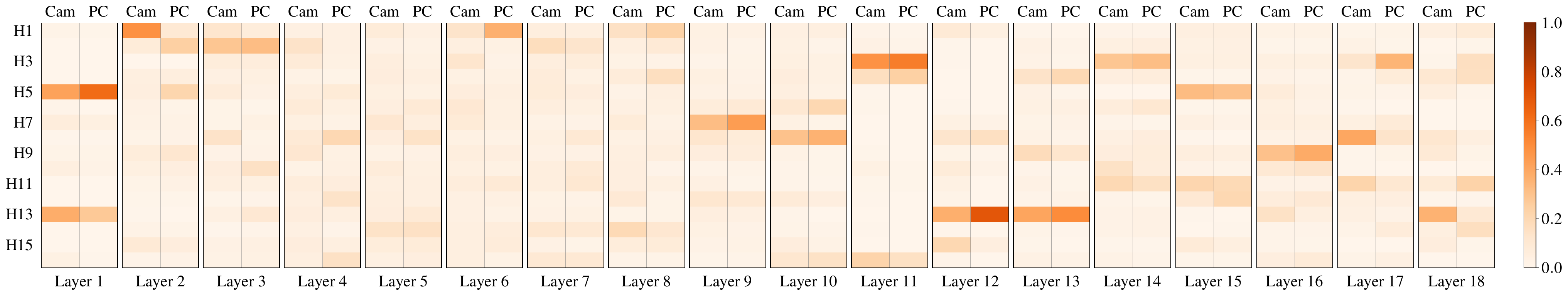}
    
    \caption{
        \textbf{HeSS Distribution in $\mathbf{\pi}^3$~\cite{wang2025pi}.}
The horizontal axis represents the Global Attention (GA) layers from GA 1 to GA 18. Each layer contains two columns, corresponding to $\mathrm{HeSS}_{\text{cam}}$ (Cam) and $\mathrm{HeSS}_{\text{pc}}$ (PC). The vertical axis lists the attention heads (H1–H16). Dark colors indicate higher sensitivity.
    }
    \label{fig:vis_scores_pi3}
\end{figure*}
\begin{figure}[t!]
    \centering
    \includegraphics[width=1\linewidth]{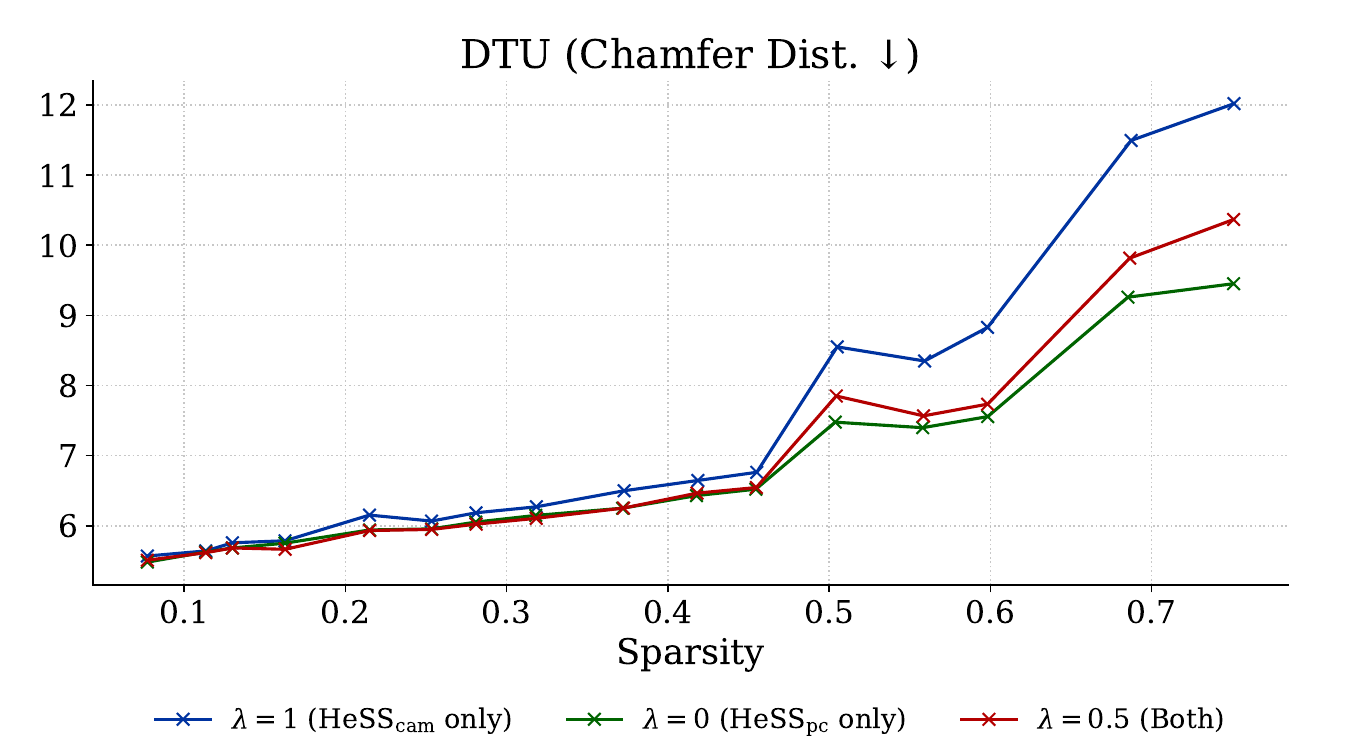}
    \caption{\textbf{Analysis of HeSS weight $\lambda$ on $\pi^3$~\cite{wang2025pi}.}
    Performance comparison across sparsity levels for $\lambda\!=\!1$ (HeSS$_{\text{cam}}$ only, \ysblue{blue}), $\lambda\!=\!0$ (HeSS$_{\text{pc}}$ only, \ysgreen{green}), and $\lambda\!=\!0.5$ (Both, \ysred{red}). 
    In contrast to VGGT~\cite{wang2025vggt}, $\pi^3$ achieves its highest accuracy with HeSS$_{\text{pc}}$ only.}
    \label{fig:hess_components_pi3}
\end{figure}

\section{Qualitative Comparison}
\begin{figure*}[t!]
    \centering
    \includegraphics[width=0.75\textwidth]{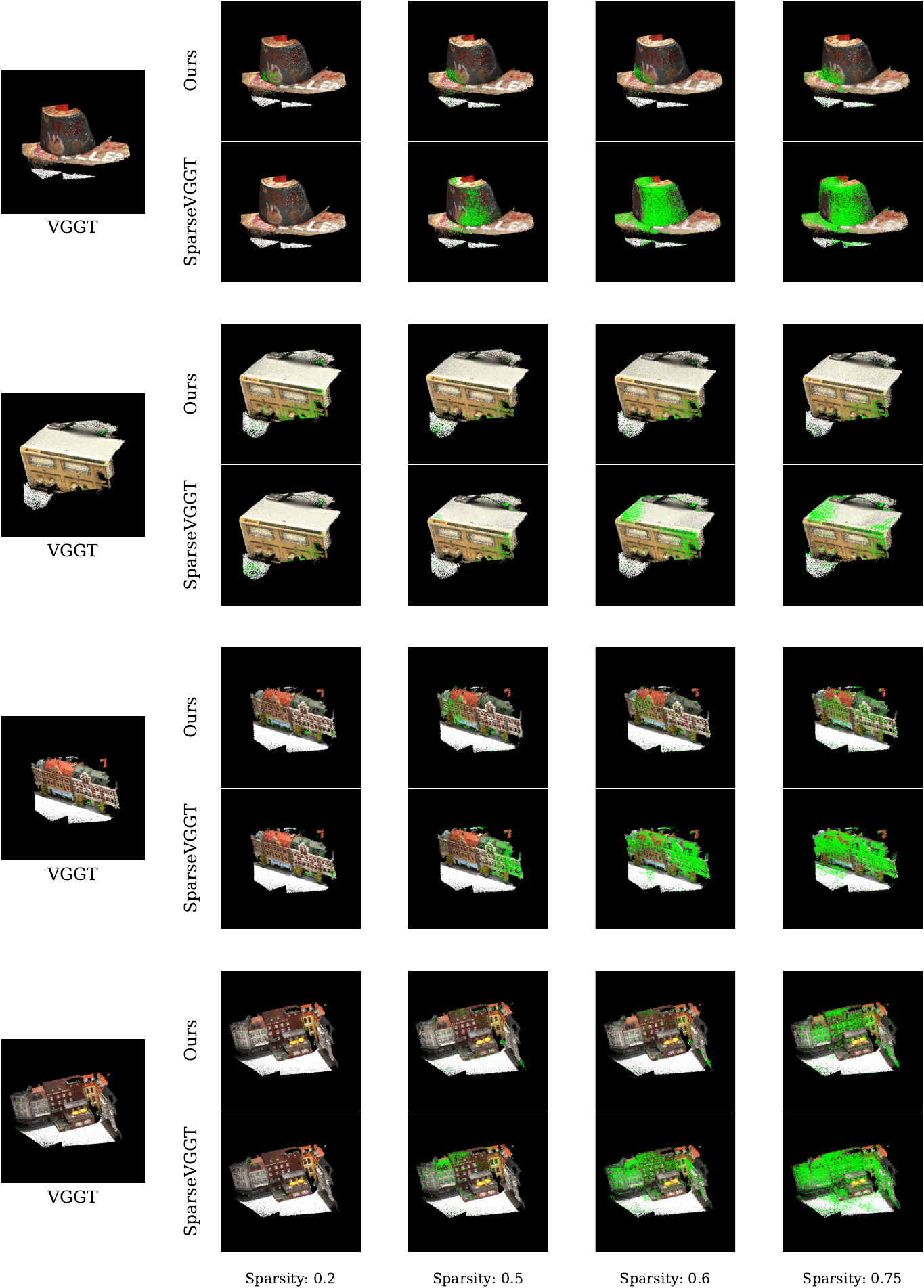}
    
    \caption{
        \textbf{Error visualization on DTU~\cite{jensen2014dtu}}
For each scene, we visualize predicted point clouds from VGGT~\cite{wang2025vggt}, SparseVGGT~\cite{wang2025fastervggt}, and our method at various sparsity levels.
Points whose 3D error exceeds 5 mm from the DTU ground truth are highlighted in green.
Our method produces fewer highlighted points than SparseVGGT, reflecting a more stable reconstruction as sparsity increases.
    }
    \label{fig:qual_res_supp_1}
\end{figure*}

\begin{figure*}[t!]
    \centering
    \includegraphics[width=0.75\textwidth]{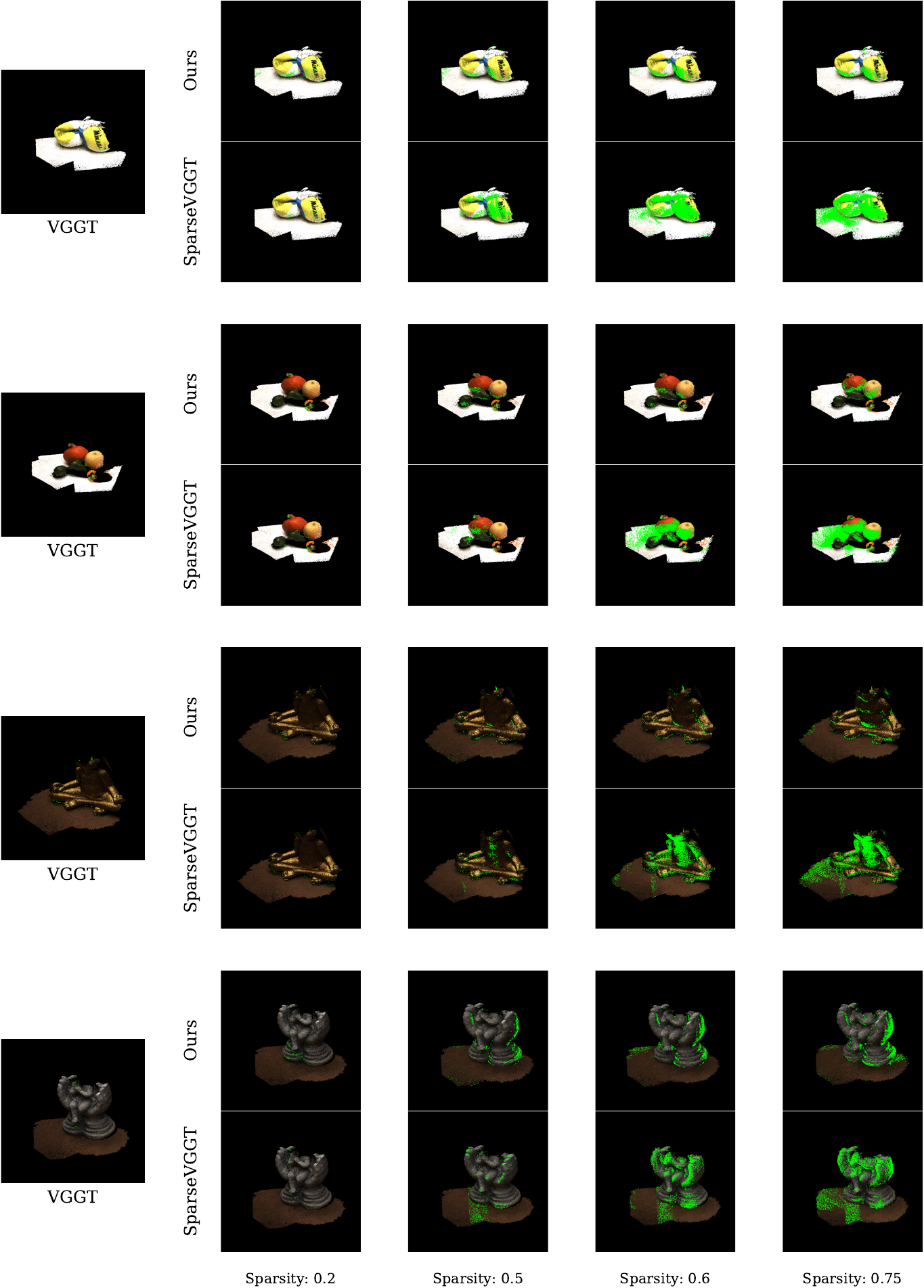}
    
    \caption{
        \textbf{Error visualization on DTU~\cite{jensen2014dtu}}
For each scene, we visualize predicted point clouds from VGGT~\cite{wang2025vggt}, SparseVGGT~\cite{wang2025fastervggt}, and our method at various sparsity levels.
Points whose 3D error exceeds 5 mm from the DTU ground truth are highlighted in green.
Our method produces fewer highlighted points than SparseVGGT, reflecting a more stable reconstruction as sparsity increases.
    }
    \label{fig:qual_res_supp_2}
\end{figure*}

For intuitive comparison, we provide qualitative results on the DTU dataset in \cref{fig:qual_res_supp_1,fig:qual_res_supp_2}.
For each predicted point cloud, we compute the nearest–neighbor distance to the ground-truth points and highlight all points whose error exceeds 5~mm in green.
Larger green regions correspond to more pronounced geometric degradation.
Compared to SparseVGGT, our method consistently produces substantially fewer green points across all sparsity levels, indicating that it effectively suppresses geometric degradation even under aggressive sparsity.

\section{Full Results Tables}
For completeness, we provide the full quantitative results on the
CO3Dv2~\cite{reizenstein2021co3d} and DTU~\cite{jensen2014dtu} datasets.
The complete tables corresponding to the main paper are reported in
\cref{tab:full_res_vggt,tab:full_res_pi}.

As sparsity increases, SparseVGGT~\cite{wang2025fastervggt} produces noticeably larger high-error regions, revealing substantial deterioration in geometric fidelity.
In contrast, our head-aware sparsification maintains stable point cloud geometry across all sparsity levels, showing only minimal growth in the highlighted high-error areas.
These observations demonstrate that our method effectively suppresses geometry degradation even under aggressive sparsity settings.

\section{Additional Experiments on $\pi^3$~\cite{wang2025pi}}
We further analyze the behavior of our method when applied to $\pi^{3}$.
This section highlights two key observations. 

\paragraph{HeSS Visualization.}
We provide HeSS visualizations for the $\pi^{3}$ model in~\cref{fig:vis_scores_pi3}. 
Similar to the distribution observed in VGGT, the sensitivity is concentrated on a subset of attention heads.

\paragraph{Effect of the Loss Weight $\lambda$.}
As shown in~\cref{fig:hess_components_pi3}, $\pi^{3}$ exhibits a markedly different trend from VGGT~\cite{wang2025vggt}
with respect to the loss weighting parameter $\lambda$.
Specifically, setting $\lambda = 0$, i.e., using only $\mathrm{HeSS}_{\text{pc}}$,
yields the best performance, whereas VGGT attains optimal results under a mixed loss with 
$\lambda > 0$.
Based on this observation, all experiments on $\pi^{3}$ adopt $\lambda = 0$.

\paragraph{Generalization to VGGT-based Architectures.}
As summarized in~\cref{tab:full_res_pi}, our method consistently outperforms SparseVGGT when applied to
$\pi^{3}$, reducing the degradation at a sparsity ratio of 0.6 by roughly
a factor of two.  
This demonstrates that HeSS-guided sparsification is not limited to the VGGT architecture
itself, but generalizes naturally to VGGT-derived models such as $\pi^{3}$,
highlighting its broader applicability across the family of VGGT-based methods.
\clearpage

\begin{table*}[t]
\centering
\small
\resizebox{0.9\textwidth}{!}{
\begin{tabular}{ccc|cc|cc|cc}
\toprule
 & & & \multicolumn{2}{c|}{DTU (Chamfer Dist.)} 
     & \multicolumn{2}{c|}{CO3Dv2 (AUC@30)} 
     & \multicolumn{2}{c}{CO3Dv2 (ATE)} \\
Sparse Ratio & CDF Thr. & Sparsity 
& SparseVGGT & \quad Ours\quad\quad 
& SparseVGGT & \quad Ours\quad\quad
& SparseVGGT & \quad Ours\quad\quad \\
\midrule
0.1 & 0.97 & 0.07 & \textbf{1.2109} & 1.2156 & 0.9545 & \textbf{0.9560} & 0.1498 & \textbf{0.1493} \\
0.3 & 0.97 & 0.10 & \textbf{1.1908} & 1.1996 & 0.9542 & \textbf{0.9558} & 0.1493 & \textbf{0.1476} \\
0.5 & 0.97 & 0.11 & 1.2618 & \textbf{1.2050} & 0.9539 & \textbf{0.9560} & 0.1520 & \textbf{0.1453} \\
0.4 & 0.95 & 0.14 & 1.2281 & \textbf{1.2219} & 0.9517 & \textbf{0.9553} & 0.1493 & \textbf{0.1444} \\
0.6 & 0.93 & 0.19 & \textbf{1.2164} & 1.2419 & 0.9482 & \textbf{0.9516} & 0.1608 & \textbf{0.1550} \\
0.5 & 0.90 & 0.23 & 1.2819 & \textbf{1.2646} & 0.9451 & \textbf{0.9504} & 0.1592 & \textbf{0.1541} \\
0.5 & 0.88 & 0.25 & \textbf{1.2751} & 1.2874 & 0.9426 & \textbf{0.9496} & 0.1640 & \textbf{0.1569} \\
0.5 & 0.85 & 0.29 & \textbf{1.2663} & 1.3389 & 0.9389 & \textbf{0.9466} & 0.1662 & \textbf{0.1593} \\
0.5 & 0.80 & 0.35 & \textbf{1.2816} & 1.4062 & 0.9327 & \textbf{0.9420} & 0.1786 & \textbf{0.1631} \\
0.5 & 0.75 & 0.39 & \textbf{1.2966} & 1.4328 & 0.9264 & \textbf{0.9396} & 0.1896 & \textbf{0.1668} \\
0.5 & 0.70 & 0.43 & \textbf{1.3188} & 1.4103 & 0.9231 & \textbf{0.9389} & 0.1870 & \textbf{0.1647} \\
0.7 & 0.70 & 0.48 & \textbf{1.3584} & 1.5011 & 0.9187 & \textbf{0.9273} & 0.1932 & \textbf{0.1815} \\
0.6 & 0.60 & 0.54 & \textbf{1.4598} & 1.4620 & 0.9118 & \textbf{0.9312} & 0.2128 & \textbf{0.1799} \\
0.6 & 0.40 & 0.60 & 1.7020 & \textbf{1.4256} & 0.8991 & \textbf{0.9283} & 0.2234 & \textbf{0.1814} \\
0.8 & 0.50 & 0.66 & 1.6800 & \textbf{1.5902} & 0.8954 & \textbf{0.9130} & 0.2407 & \textbf{0.2176} \\
0.8 & 0.40 & 0.73 & 1.9287 & \textbf{1.6030} & 0.8866 & \textbf{0.9115} & 0.2527 & \textbf{0.2184} \\
\bottomrule\textbf{}
\end{tabular}
}
\caption{
\textbf{Comparison of SparseVGGT~\cite{wang2025fastervggt} and our method with the VGGT~\cite{wang2025vggt} backbone.}
Performance is evaluated on DTU~\cite{jensen2014dtu} (Chamfer Distance) and CO3Dv2~\cite{reizenstein2021co3d} (AUC@30, ATE) across varying sparsity levels.
}
\label{tab:full_res_vggt}
\end{table*}
\begin{table*}[t]
\centering
\small
\resizebox{0.9\textwidth}{!}{
\begin{tabular}{ccc|cc|cc|cc}
\toprule
 & & & \multicolumn{2}{c|}{DTU (Chamfer Dist.)} 
     & \multicolumn{2}{c|}{CO3Dv2 (AUC@30)} 
     & \multicolumn{2}{c}{CO3Dv2 (ATE)} \\
Sparse Ratio & CDF Thr. & Sparsity 
& SparseVGGT & \quad Ours\quad\quad 
& SparseVGGT & \quad Ours\quad\quad
& SparseVGGT & \quad Ours\quad\quad \\
\midrule
0.1 & 0.97 & 0.08 & 1.3320 & \textbf{1.3151} & 0.9722 & \textbf{0.9728} & 5.581  & \textbf{5.4822} \\
0.3 & 0.97 & 0.11 & 1.3515 & \textbf{1.3217} & 0.9713 & \textbf{0.9726} & 5.7365 & \textbf{5.6254} \\
0.5 & 0.97 & 0.13 & 1.3536 & \textbf{1.3286} & 0.9709 & \textbf{0.9716} & 5.8080 & \textbf{5.6833} \\
0.4 & 0.95 & 0.16 & 1.3680 & \textbf{1.3260} & 0.9704 & \textbf{0.9722} & 5.9064 & \textbf{5.7517} \\
0.6 & 0.93 & 0.22 & 1.3962 & \textbf{1.3559} & 0.9688 & \textbf{0.9708} & 6.1670 & \textbf{5.9403} \\
0.5 & 0.90 & 0.25 & 1.4450 & \textbf{1.3479} & 0.9683 & \textbf{0.9708} & 6.2365 & \textbf{5.9556} \\
0.5 & 0.88 & 0.28 & 1.4759 & \textbf{1.3536} & 0.9673 & \textbf{0.9712} & 6.4240 & \textbf{6.0526} \\
0.5 & 0.85 & 0.32 & 1.5778 & \textbf{1.3645} & 0.9669 & \textbf{0.9699} & 6.5726 & \textbf{6.1476} \\
0.5 & 0.80 & 0.37 & 1.7098 & \textbf{1.3932} & 0.9652 & \textbf{0.9700} & 6.8560 & \textbf{6.2508} \\
0.5 & 0.75 & 0.42 & 1.8843 & \textbf{1.4426} & 0.9646 & \textbf{0.9701} & 7.0518 & \textbf{6.4302} \\
0.5 & 0.70 & 0.46 & 2.0860 & \textbf{1.4315} & 0.9627 & \textbf{0.9690} & 7.3248 & \textbf{6.5217} \\
0.7 & 0.70 & 0.51 & 2.0763 & \textbf{1.5375} & 0.9608 & \textbf{0.9631} & 7.6261 & \textbf{7.4775} \\
0.6 & 0.60 & 0.56 & 2.4659 & \textbf{1.8541} & 0.9596 & \textbf{0.9635} & 7.9918 & \textbf{7.3988} \\
0.6 & 0.40 & 0.60 & 2.8635 & \textbf{2.0883} & 0.9573 & \textbf{0.9627} & 8.3667 & \textbf{7.5568} \\
0.8 & 0.50 & 0.69 & 2.7606 & \textbf{2.3020} & 0.9495 & \textbf{0.9505} & 9.5964 & \textbf{9.2602} \\
0.8 & 0.40 & 0.75 & 3.0046 & \textbf{2.5796} & \textbf{0.9458} & 0.9455 & 10.1408 & \textbf{9.4527} \\
\bottomrule
\end{tabular}
}
\caption{\textbf{Comparison of SparseVGGT~\cite{wang2025fastervggt} and our method with the $\pi^3$~\cite{wang2025pi} backbone.} 
Performance is evaluated on DTU~\cite{jensen2014dtu} (Chamfer Distance) and CO3Dv2~\cite{reizenstein2021co3d} (AUC@30, ATE) across varying sparsity levels.}
\label{tab:full_res_pi}
\end{table*}
\clearpage


\end{document}